%% file: acl_latex.tex
\documentclass[11pt]{article}

\PassOptionsToPackage{table,dvipsnames,svgnames}{xcolor}

\usepackage[final]{acl}

\usepackage{times}
\usepackage{latexsym}

\usepackage[T1]{fontenc}

\usepackage[utf8]{inputenc}

\usepackage{microtype}

\usepackage{inconsolata}

\usepackage{graphicx}
\usepackage{fontawesome5}
\usepackage{microtype}
\usepackage{hyperref}
\usepackage{url}
\usepackage{booktabs}
\usepackage{amsmath,amssymb}
\usepackage{dsfont}                
\providecommand{\indic}{\mathds{1}}
\usepackage{amsthm}

\usepackage{enumitem}
\usepackage{tcolorbox}
\usepackage{algorithm}
\usepackage{algpseudocode}
\usepackage{tikz}
\usetikzlibrary{arrows.meta, positioning, calc, decorations.pathmorphing,
                shapes.geometric, fit, backgrounds}

\usepackage{iftex}
\ifPDFTeX
  \providecommand{\excalifont}{\sffamily\bfseries}
\else
  \usepackage{fontspec}
  \IfFontExistsTF{[fig/fonts/Excalifont-Regular.ttf]}{%
    \newfontfamily\excalifont{Excalifont-Regular.ttf}[
      Path=fig/fonts/, AutoFakeBold=2.5, AutoFakeSlant=0.18]%
  }{%
    \providecommand{\excalifont}{\sffamily\bfseries}%
  }
\fi

\usepackage{graphicx}       
\usepackage{subcaption}   
\usepackage{wrapfig}
\usepackage{verbatim}
\usepackage[table, dvipsnames, svgnames]{xcolor}
\usepackage{bbding}
\usepackage{pifont}
\tcbuselibrary{skins, listings, breakable}
\usepackage{fancyvrb} 
\usepackage{fvextra} 
\usepackage{listings}
\usepackage{titletoc}
\usepackage[toc,page,header]{appendix}
\usepackage{minitoc}
\usepackage{scalerel}
\usepackage{cuted}
\definecolor{darkblue}{rgb}{0, 0, 0.5}
\definecolor{darkgreen}{RGB}{0, 100, 0}
\hypersetup{colorlinks=true, citecolor=darkblue, linkcolor=darkblue, urlcolor=darkblue}

\DefineVerbatimEnvironment{MyBox}{Verbatim}{
frame=single, 
framesep=6pt, 
fontsize=\small, 
breaklines=true, 
breakanywhere=true, 
breaksymbol={}, 
breaksymbolsepleft=0pt, 
breaksymbolsepright=0pt, 
}

\newtcolorbox{case}[1][]{
  colback=black!5,   
  colframe=black!75, 
  boxrule=1pt,       
  arc=4pt,           
  left=6pt, right=6pt, top=6pt, bottom=6pt,
  breakable,               
  #1
}

\newcommand{\huggingfaceicon}{\scalebox{1.5}{\scalerel*{\includegraphics{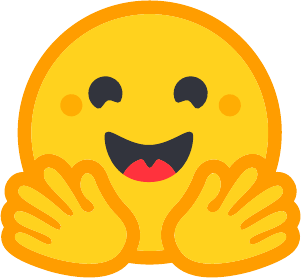}}{X}}}

\newcommand{\microsofticon}{\scalerel*{\includegraphics{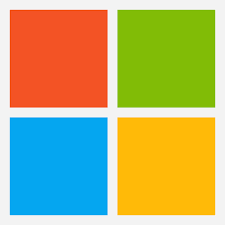}}{X}}

\newcommand{\microsoftAIicon}{\scalerel*{\includegraphics{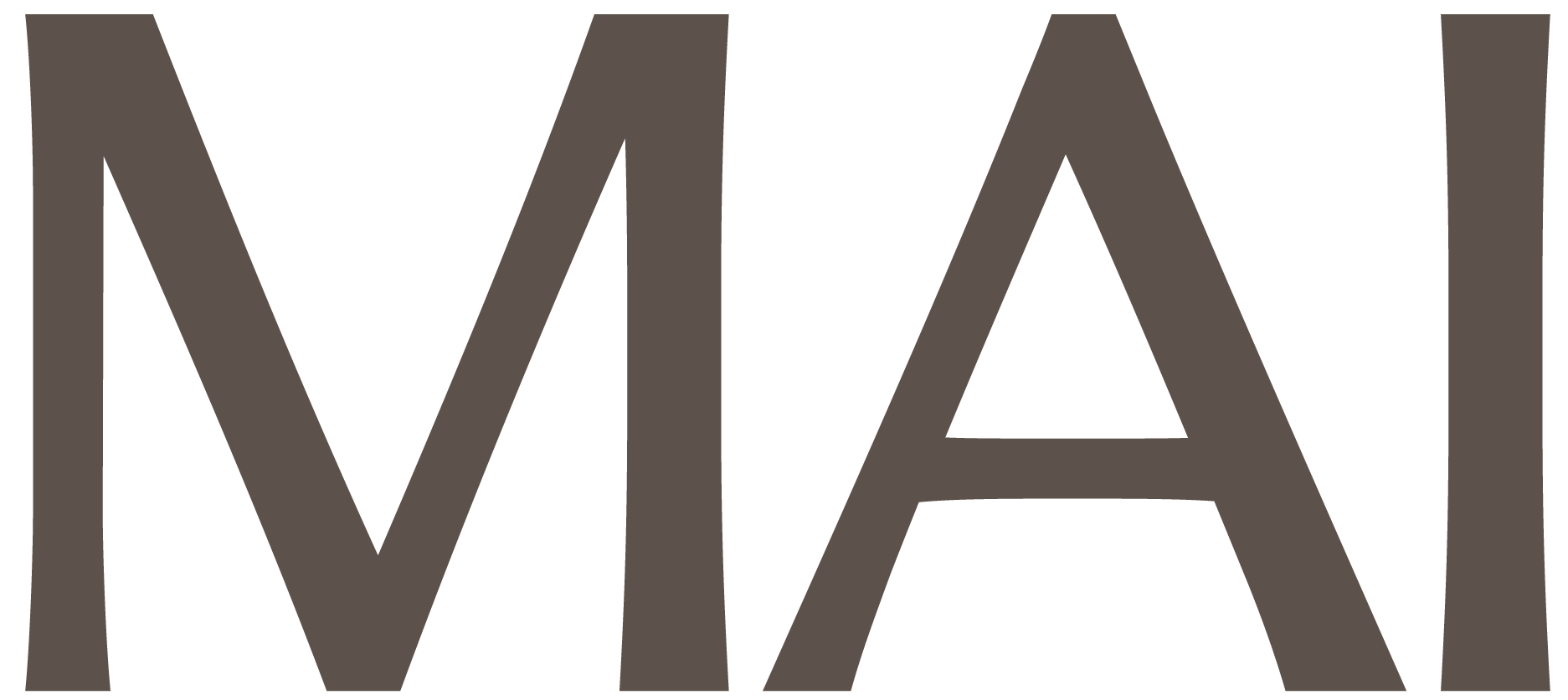}}{X}}

\definecolor{deepred}{RGB}{180,0,0}
\definecolor{harvestgold}{rgb}{0.85, 0.57, 0.0}

\definecolor{deepred}{RGB}{180,0,0}
\definecolor{pathgreen}{RGB}{0,130,72}

\DeclareRobustCommand{\caprtitlelogo}{%
  \csname tikz\endcsname[baseline=-0.48ex, x=1em, y=1em, line cap=round, line join=round]{%
    \path[use as bounding box] (-1.25,-0.70) rectangle (1.47,0.61);
    \coordinate (pivot) at (-0.18,0.00);
    \draw[RoyalBlue!82!black, line width=1.12pt]
      (-0.83,0.00) .. controls (-0.62,0.24) and (-0.40,-0.05) .. (pivot);
    \draw[deepred!86!black, line width=0.84pt, dash pattern=on 1.10pt off 1.35pt]
      (pivot) .. controls (0.18,0.39) and (0.62,0.56) .. (1.02,0.50);
    \draw[pathgreen!86!black, line width=1.08pt]
      (pivot) .. controls (0.20,-0.43) and (0.60,-0.57) .. (0.97,-0.50);

    \fill[RoyalBlue!10, rounded corners=0.035em] (-1.24,-0.22) rectangle (-0.86,0.22);
    \draw[RoyalBlue!78!black, line width=0.46pt, rounded corners=0.035em]
      (-1.24,-0.22) rectangle (-0.86,0.22);
    \foreach \x/\y/\c in {-1.17/0.15/RoyalBlue!36,-0.99/0.15/RoyalBlue!18,
                          -1.17/-0.15/RoyalBlue!16,-0.99/-0.15/RoyalBlue!34}{%
      \fill[\c] (\x-0.042,\y-0.042) rectangle (\x+0.042,\y+0.042);
    }%
    \fill[black!70] (-1.13,0.055) circle (0.018);
    \fill[black!70] (-0.99,0.055) circle (0.018);
    \draw[black!70, line width=0.28pt]
      (-1.13,-0.070) .. controls (-1.08,-0.030) and (-1.04,-0.110) .. (-0.98,-0.070);

    \fill[white, rounded corners=0.035em] (-0.30,-0.115) rectangle (-0.06,0.115);
    \draw[RoyalBlue!82!black, line width=0.52pt, rounded corners=0.035em]
      (-0.30,-0.115) rectangle (-0.06,0.115);
    \draw[RoyalBlue!55!black, line width=0.28pt] (-0.25,0.052) -- (-0.11,0.052);
    \draw[RoyalBlue!55!black, line width=0.28pt] (-0.25,0.000) -- (-0.11,0.000);
    \draw[RoyalBlue!55!black, line width=0.28pt] (-0.25,-0.052) -- (-0.11,-0.052);

    \draw[deepred!86!black, line width=1.02pt]
      (1.16,0.38) -- (1.35,0.57) (1.35,0.38) -- (1.16,0.57);

    \foreach \x in {1.08,1.34}{%
      \fill[pathgreen!12, rounded corners=0.030em] (\x-0.115,-0.685) rectangle (\x+0.115,-0.435);
      \draw[pathgreen!78!black, line width=0.42pt, rounded corners=0.030em]
        (\x-0.115,-0.685) rectangle (\x+0.115,-0.435);
      \fill[black!70] (\x-0.043,-0.532) circle (0.011);
      \fill[black!70] (\x+0.043,-0.532) circle (0.011);
      \draw[black!70, line width=0.24pt]
        (\x-0.052,-0.600) .. controls (\x-0.018,-0.640) and (\x+0.018,-0.640) .. (\x+0.052,-0.600);
    }%
  }%
}

\title{\texorpdfstring{\makebox[0pt][r]{\raisebox{-0.4\height}[0pt][0pt]{\rotatebox[origin=c]{270}{\scalebox{1.50}{\caprtitlelogo}}}\hspace{1.2em}}}{}Read the Trace, Steer the Path: Trajectory-Aware\\Reinforcement Learning for Diffusion Language Models}

\author{\textbf{Anant Khandelwal}, \textbf{Manish Gupta} \\
  \microsofticon\; \microsoftAIicon\; Microsoft AI, India\\
  \texttt{\{anantk,gmanish\}@microsoft.com} \\
  \small
\faGithub\; \url{https://github.com/infusion-zero-edit/CAPR}\\
\small
\huggingfaceicon\; \url{https://huggingface.co/collections/anantk/capr-checkpoints}
}

\begin{document}
\maketitle
\begin{abstract}
Diffusion large language models (dLLMs) generate responses by iteratively unmasking and revising many positions in parallel. This process leaves a rich denoising trace depicting which tokens become confident, which remain unstable, and when commitments form. Existing dLLM reinforcement learning methods use this signal only weakly. Flat rollouts are cheap, but assign a single outcome reward to the whole trajectory. Tree rollouts provide finer, verifiable training signals by branching partial trajectories and propagating leaf rewards upward, but is compute intensive. We ask whether the denoising trace itself can provide tree-like supervision without tree-level compute. We introduce CAPR (\textbf{C}ached-\textbf{A}mortized \textbf{P}ath \textbf{R}efinement), a dLLM-RL algorithm that summarizes the denoising trace into a compact path state, uses cached trajectory states to generate cheap sibling continuations, and trains a block-level value head for local block-wise supervision. Under a block-wise unmasking schedule, CAPR records path-state and block-progress features, then redistributes the final outcome reward across blocks according to the tokens revealed in each block. This trains the value head to convert one sparse reward into block-level PPO weights. CAPR therefore recovers much of the granularity of tree search while avoiding full tree expansion, reducing rollout-generation cost to roughly 0.75$\times$ of flat rollouts and 0.6$\times$ of tree rollouts (at standard setting). Across 4×4 Sudoku, Countdown, GSM8K, and Math500, on dense and mixture-of-experts LLaDA backbones, CAPR sets a new state of the art for RL-tuned dLLMs at 256- and 512-token budgets. On Sudoku, it matches the strongest tree-structured baseline at less than one third of the per-step compute.
\end{abstract}

\section{Introduction}
\label{sec:intro}
Diffusion large language models (dLLMs)~\citep{nie2025llada,zhu2025lladamoe} generate text by denoising a fully masked sequence over many parallel steps. At every step, the model exposes a predictive distribution for every masked position. This is a rich trace of how the answer forms, but current reinforcement learning (RL) based post-training for dLLMs~\citep{zhao2025d1,zhu2025llada1_5,tang2025wd1,xie2025step,pan2025dtreerpo} uses little of it. An end-of-generation reward must still be assigned across many denoising steps and token blocks, so the learner has no direct signal for which step fixed an important token and which step merely changed an irrelevant one.
\begin{figure*}[!t]
\centering
\includegraphics[width=0.7\linewidth]{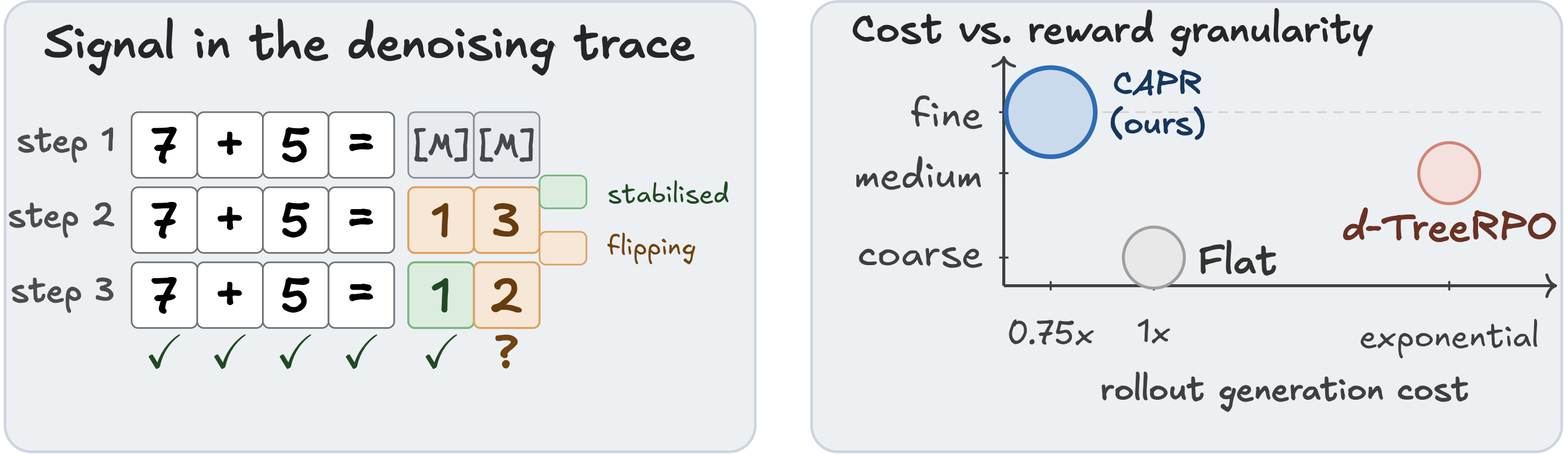}
\caption{The denoising trace contains information that current RL methods do not use. Left: each masked position has a sequence of provisional predictions however stable and confident positions can be separated from positions that keep changing. Right: prior dLLM-RL methods occupy two extremes of the cost--granularity plane. Flat rollouts are cheap but give one scalar advantage per trajectory, \emph{d}-TreeRPO~\citep{pan2025dtreerpo} pays tree-expansion cost for a small number of segment-level advantages. The approach introduced in this paper named, CAPR, uses the trace to obtain per-block advantages at roughly $0.75\times$ flat-rollout cost.}
\label{fig:teaser}
\end{figure*}

Existing RL methods for dLLMs sit at two extremes. Flat-rollout methods assign the same outcome reward to all steps. They are cheap, but they cannot tell which denoising decisions mattered. Tree-rollout methods~\citep{pan2025dtreerpo} fork alternative continuations and compare their final rewards. They obtain finer advantages, but their cost grows quickly with branching factor and depth. Figure~\ref{fig:teaser} (right) shows this cost-granularity tradeoff. The denoising trace offers a third route. At each step, a position either becomes more stable and confident, or it keeps changing. Figure~\ref{fig:teaser} (left) illustrates this on a simple prompt (``$7{+}5$:''). By step three the tens digit has settled at \texttt{1}, while the units digit is still oscillating. The trace therefore tells us which positions are becoming committed, without running a tree search. The key question is how to convert this trace into a learning signal. 

We answer this question with \textbf{CAPR}(\textbf{C}ached-\textbf{A}mortized \textbf{P}ath \textbf{R}efinement), an RL algorithm for block-based dLLMs. CAPR consists of three broad mechanisms. a) \emph{Cache \&\ Steer}: CAPR records the early denoising steps, builds a compact path state from confidence and stability at each position, and reuses this cached prefix for producing sibling rollouts or branches. The path state is used immediately at every denoising step to steer the next-step logits, reinforcing stable commitments and damping oscillating ones. b) \emph{Branch \&\ Prune}: At a fixed mid-trajectory step, CAPR branches a small number of Gumbel-perturbed sibling continuations (rollouts when completed) from the cached prefix. At block boundaries, it keeps the siblings whose path states indicate the strongest commitments while pruning others. Finally, a small c) \emph{Block Critic} that reads the path state at each block boundary and converts the outcome reward into one advantage per block for the PPO update. Figure~\ref{fig:method} depicts the three mechanisms. This design keeps the useful part of tree methods which makes local comparison among alternatives, without expanding a full tree. 

Due to caching and pruning, CAPR is much more efficient compared to both flat rollouts as well as tree rollouts. Figure~\ref{fig:teaser}(right) shows that CAPR consumes 0.75$\times$ of the flat rollout generation cost. At standard d-TreeRPO tree rollout settings of branching factor$=4$ and tree depth$=2$, CAPR incurs 0.6$\times$ of the tree rollout generation cost as shown in Table~\ref{tab:complexity} in Appendix~\ref{app:complexity}.

CAPR is not just efficient but also more accurate compared to both flat and tree rollouts. On four reasoning benchmarks ($4{\times}4$ Sudoku, Countdown, GSM8K, Math500) and two open-source LLaDA backbones spanning dense and mixture-of-experts variants, CAPR sets a new state of the art for RL-tuned dLLMs at both the 256-token and 512-token decoding budgets. On Sudoku, it matches the strongest tree-structured baseline at less than a third of the per-step compute (Table~\ref{tab:compute}). Ablations show that all three mechanisms help, with per-block advantages providing the largest gain.

\paragraph{Contributions.}
\begin{itemize}
[leftmargin=1.2em,topsep=2pt,itemsep=0pt]
\item A denoising trace-based view of RL for masked dLLMs: confidence and stability along the denoising path can be turned into per-block advantages, without full tree expansion.
\item CAPR, a block-based RL algorithm that caches early denoising outputs, amortizes sibling rollouts from the cached prefix, steers each next denoising step with a path state, and uses a block critic to turn the final reward into one advantage per block for PPO.
\item State-of-the-art results for RL-tuned dLLMs at both 256-token and 512-token decoding budgets across four reasoning benchmarks and two LLaDA backbones, with less than a third of the per-step compute of the strongest tree-structured baseline on Sudoku.
\end{itemize}


\section{Related Work}
\label{sec:related}
\paragraph{Diffusion language models.}
Diffusion language models generate text by iteratively denoising masked tokens rather than predicting strictly left to right~\citep{nie2025llada,zhu2025lladamoe,sahoo2024simple,shi2024simplified,nie2024scaling,ye2023diffusion,ye2025dream,khanna2025mercury,song2025seed}. This decoding process exposes a distribution over every masked position at every step, which makes dLLMs a natural setting for trace-based training signals. Recent open models and paradigms, including LLaDA-1.5~\citep{zhu2025llada1_5}, LLaDA-MoE~\citep{zhu2025lladamoe}, Dream-7B~\citep{ye2025dream}, and SDAR~\citep{cheng2025sdar}, have made dLLMs strong reasoning backbones. We use the dense and MoE LLaDA variants in our experiments.

\paragraph{RL post-training for dLLMs.}
Most RL methods for dLLMs differ in how they assign an outcome reward to the denoising process. Flat-rollout methods, including Diffu-GRPO~\citep{zhao2025d1}, VRPO~\citep{zhu2025llada1_5}, wd1~\citep{tang2025wd1}, and d2-stepMerge~\citep{wang2025d2}, keep sampling cheap by applying one trajectory-level reward to all token updates. This makes the update simple, but it does not identify which denoising steps or blocks mattered. Step-aware methods such as SAPO~\citep{xie2025step} and TraceRL~\citep{wang2025revolutionizing} attach more local signals inside a rollout, but they do not use the denoising trace to learn a block-level value function. Tree-based methods, especially \emph{d}-TreeRPO~\citep{pan2025dtreerpo}, compare branched continuations and produce segment-level advantages. They are effective, but the cost grows rapidly with branching factor and depth. CAPR targets the gap between these regimes. It keeps the rollout structure close to flat sampling, but uses the trace to produce one advantage per block.

Concurrent work explores related ways to stabilize or enrich dLLM-RL, including consistency-trajectory regularization~\citep{yang2025taming}, lateral-thought reinforcement~\citep{huang2025reinforcing}, multimodal RL for dLLMs~\citep{yang2025mmada}, entropy and anchoring objectives~\citep{yang2025entropic,cui2025entropy,li2025entropy}, boundary-aware updates~\citep{lin2025boundary}, MDPO~\citep{he2025mdpo}, principled denoising objectives~\citep{ou2025principled}, and inpainting-style RL~\citep{zhao2025inpainting}. CAPR is complementary to these directions. Its main contribution is to convert the denoising trace into reusable path states and per-block PPO weights.

\paragraph{Connections to policy optimization and guided decoding.}
CAPR's block critic is a lightweight actor--critic component~\citep{sutton1988learning,schulman2015high,schulman2017proximal} adapted to sparse, outcome-only rewards. It reads features that are specific to masked decoding: path-state statistics, block progress, prompt-level reward statistics, and the tokens revealed so far. A slowly-updated target copy~\citep{lillicrap2015continuous} and a small replay buffer keep this online value head stable.

The steering shift is related to guided decoding, but it does not use an external classifier or reward model. It uses the model's own trace. Stable, confident positions receive a small positive bias on the next denoising step, while oscillating positions are damped. The form follows the additive logit control induced by linearly-solvable MDPs~\citep{todorov2006linearly}. Finally, CAPR uses a one-sided $\chi^2$ trust region
to suppress rare token-level log-ratio spikes without penalizing all policy movement equally. This is related in spirit to asymmetric clipping~\citep{yu2025dapo,xie2025step} and policy-gradient stabilization~\citep{cui2025entropy,zhang2024entropy,tan2025gtpo,zheng2025group}, but it is applied to token-level dLLM updates with an explicit upper-tail trust-region interpretation.

\section{Method}
\label{sec:method}

\subsection{Preliminaries}
\label{sec:formulation}
A masked dLLM $\pi_\theta$ generates a completion $y$ for prompt $q$ over $N$ denoising steps, moving from $z_N=[q;\,\texttt{[MASK]}^L]$ to $z_0=[q;\,y]$. Block decoding partitions the $L$ completion tokens into blocks. Flat GRPO~\citep{shao2024deepseekmath} optimizes a clipped policy-gradient objective with a rollout-level outcome advantage shared by all tokens in the completion. Tree methods~\citep{pan2025dtreerpo} obtain more local advantages by comparing branched continuations. CAPR keeps the sampling pattern close to flat GRPO, but replaces the single rollout-level advantage with one advantage per block. We write $r_{ij}(\theta)=\pi_\theta(y_{ij})/\pi_{\theta_{\rm old}}(y_{ij})$ for the token-level importance ratio used by the PPO update in CAPR.

\begin{figure*}[t]
    \centering
    \makebox[\linewidth][c]{\includegraphics[width=0.8\linewidth]{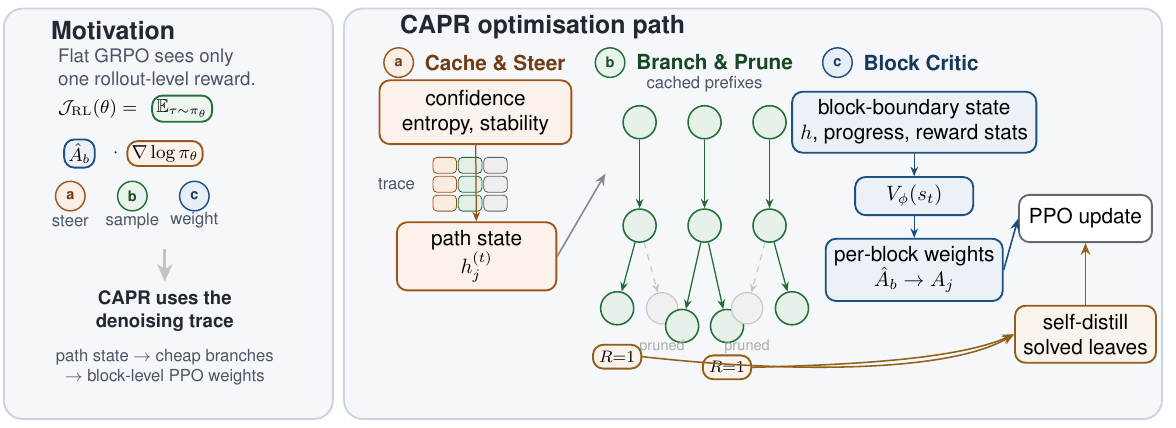}}
    \caption{Overview of CAPR. \textbf{(a) Trace State / Cache \& Steer}: CAPR summarizes per-position confidence, entropy, and stability into a path state, then uses it to steer the next reverse step while carrying only the previous clean-token prediction and path state. \textbf{(b) Branch \& Prune}: a shared denoising prefix is forked once at the branch step, and siblings are kept by path-state quality. \textbf{(c) Block Critic}: a value head reads block-boundary states and converts the final reward into per-block PPO weights. Self-distillation uses solved rollouts as supervised targets so zero-variance groups still provide learning signal.}
    \label{fig:method}
\end{figure*}

The CAPR methodology is illustrated in Fig.~\ref{fig:method}. During the early denoising steps it records how each masked position behaves across steps. Stable, confident positions are treated as emerging commitments, oscillating positions are treated as unresolved. This record is compressed into a path state. The path state is then used in three ways: it steers the transition $z_t$ to $z_{t-1}$, later branches reuse a cached prefix, and it gives the block critic the information needed to turn one final reward into one advantage per block.

\subsection{\textsc{Cache \& Steer}}
\label{sec:controller}

At each denoising step $t$, CAPR reads how each position $j$ is behaving by recording confidence in the current top token, entropy of the probability distribution per position at the current step, and whether the top token stayed stable across denoising steps. These are packed into a small feature vector $\phi_j^{(t)} = \big(c_j^{(t)}, s_j^{(t)}, e_j^{(t)}, a_j^{(t)}\big)$ where $c_j$ is confidence, $s_j$ is top-token stability, $e_j$ is normalized entropy, and $a_j$ is certainty. The feature computations are defined in App.~\ref{app:metastate}. 

The path state $h_j^{(t)}$ is updated in GRU-style~\citep{cho2014learning}. A reset gate forgets stale evidence when entropy is high, and an update gate preserves memory when the token has remained stable and certain. Thus $h_j$ acts as a signed memory of commitment. Positive $h_j$ means the position is stable, negative $h_j$ means the model is still oscillating. The exact analytic gates are defined in App.~\ref{app:metastate}.
\begin{align}
h_j^{(t)} &\leftarrow \mathrm{PathGRU}\big(h_j^{(t-1)},\phi_j^{(t)}\big),
\label{eq:pathgru}
\end{align}

At step $t$, CAPR uses $z_t$ to steer the resulting logits and then identifies the positions and sampled token predictions to form $z_{t-1}$. 
Across steps, only $z_t$ and the path state $h_j^{(t)}$ are carried over and the logits themselves are recomputed each step. 
CAPR steers the logits as follows. 
\begin{equation}
\ell_{j}^{(t)} \mathrel{+}= \frac{h_j^{(t)}\, c_j^{(t)}\,(2 s_j^{(t)}-1)}{\lambda}
\label{eq:lsmdp}
\end{equation}
where $\lambda$ is a path state steering scaling constant. A stable and confident token receives a positive push and an oscillating token is dampened. 

The same path state adjusts the confidence used to choose which masked positions are revealed as well as which unmasked positions should be remasked. First, we compute $\kappa_j^{(t)}$ as follows.
\begin{equation}
\kappa_j^{(t)} = c_j^{(t)} + \gamma h_j^{(t)} a_j^{(t)}\frac{1+s_j^{(t)}}{2},
\label{eq:metastate_injector}
\end{equation}
where $\gamma$ is a path state injection scaling constant.

$\kappa_j$ steers which positions are trusted enough to unmask. If the denoising schedule decides to unmask $K_t$ tokens, CAPR selects
\begin{equation}
\mathcal{U}_t = \operatorname{TopK}_{j:z_{t,j}=\texttt{[MASK]}}\big(\kappa_j^{(t)},K_t\big),
\label{eq:metastate_unmask}
\end{equation}

For these positions in $\mathcal{U}_t$ identified for unmasking, the logit bias steers which token is sampled. After sampling, the new prediction replaces $z_t$ for the following denoising step and the logits themselves are not carried forward. Both steering signals in Eq.~\ref{eq:metastate_injector} and~\ref{eq:metastate_unmask} come from the model's own denoising trace rather than an external classifier.

\subsection{\textsc{Branch \& Prune}}
\label{sec:branching}

To explore alternative completions (see App.~\ref{app:proof}), and to assign granular rewards for every block in every branch, at low compute cost, we perform branching at the midpoint of the denoising trajectory in CAPR. To reduce compute costs for each branch, CAPR caches a prefix so that these sibling continuations (i.e. branches) can start without any extra forward pass. The prefix is a combination of logits distribution at the branch step, along with the predicted tokens, and path state $h^{(t)}$. 

At a fixed fraction $f_b$ (we set it to 0.5) of the denoising budget $N$, it forks $G_b$ branches from the shared prefix. The fork reuses the stored branch-step prefix and changes only the predictions for the $G_b$ branches by sampling a different $\mathcal{G}_v$ for every vocab token $v$ in each branch. The predictions for each branch are then calculated as 
$\arg\max_v\big(\ell_{j,v}^{(t)}/\tau + \mathcal{G}_v\big)$ for every position $j$ in $\mathcal{U}_t$ at $t=f_b\times N$ 
with $\mathcal{G}_v \sim \mathrm{Gumbel}(0,1)$. $\tau$ is the sampling temperature. Each sibling copies the current path state and then completes the remaining denoising steps independently.

This prefix caching gives the compute advantage. One branch group costs
$\big[f_b + G_b(1-f_b)\big]N$
forward steps, detailed comparision with baselines on forward cost is given in App.~\ref{app:complexity}. At $G_b$=2, and $f_b$=0.5, cost per rollout is $\frac{1}{G_b}\big[f_b + G_b(1-f_b)\big]N$=0.75$N$. To produce $G$ rollouts, we run $M=\lceil G/G_b\rceil$ such groups. 

At a regular frequency schedule after the fork, CAPR keeps the most useful branches rather than carrying all of them forward. The score is computed based on mean path state $h^{(t)}$ of the branch and hence branches with more positions settled into stable predictions receive higher scores. This pruning is deliberately local and cheap. It does not require evaluating a reward model at intermediate steps, it uses the same trace signal already maintained by ``Cache \& Steer''.

Tree method branches many times and compares final rewards of many continuations, but this expands compute quickly. CAPR instead creates a small local comparison set by forking sibling continuations from a cached mid-trajectory state. Since the siblings share the same prefix and differ only in the suffix, their outcomes and path-state quality give useful evidence about which later denoising decisions are promising. Pruning them prevents the small branch set from growing into a full tree.

\subsection{\textsc{Block Critic}}
\label{sec:blockcritic}
The block critic answers a simple question: after seeing only the final reward, which blocks were most useful? At each block boundary, CAPR stores a compact state containing the path state, block progress (number of unmasked tokens within the block), prompt-level reward statistics, and tokens revealed so far. Using this compact state, the value head is trained as discussed in App.~\ref{app:blockcritic}.

After the rollout finishes, CAPR redistributes the final reward $R$ over blocks according to how many tokens each block revealed, with the final block also receiving the terminal outcome signal. TD/GAE~\citep{schulman2015high,sutton1988learning} with a slowly updated target head $V_{\bar\phi}$~\citep{lillicrap2015continuous} then produces one block advantage $\hat A_b$ for each block $b$. Thus, the value head converts one sparse outcome reward into block-level weights over the denoising path.

For PPO, these block advantages are used only to reweight the rollout-level GRPO advantage. Let $b_i(j)$ denote the block in which token $j$ of rollout $i$ is revealed. CAPR computes
\begin{equation}
\begin{aligned}
w_{i,b} &= B\,\mathrm{softmax}(\hat A_{i,1:B})_b, \\
\widetilde A_{i,j}
&= A_i^{\rm GRPO}
\big[(1-\beta_{\rm TD})
+\beta_{\rm TD}\,w_{i,b_i(j)}\big],
\end{aligned}
\label{eq:tokenadv}
\end{equation}
where $B$ is the number of blocks. The factor $B$ keeps the mean block weight near one, so the block critic redistributes the rollout-level signal rather than changing its overall scale. Every token revealed in the same block receives the same block weight. When the value head is poorly calibrated, the blend $\beta_{\rm TD}$ is reduced and PPO falls back toward the flat GRPO advantage. Appendix~\ref{app:blockcritic} gives the value-head features, TD targets, and blending gate. At inference, the value head is removed and only the trained LoRA-adapted policy remains.

\subsection{Training Objective and CAPR Algorithm}
\label{sec:objective}

The complete CAPR objective is
\begin{equation}
\begin{aligned}
\mathcal{L}_{\rm CAPR} &= \mathcal{L}_{\rm PPO} + \beta\,\mathcal{L}_{\rm KL} + \lambda_f\mathcal{L}_{\chi^2} \\
&\quad + \mathcal{L}_{\rm NLL} + c_v\mathcal{L}_V.
\end{aligned}
\label{eq:total}
\end{equation}
In Eq.~\ref{eq:total}, $\mathcal{L}_{\rm PPO}$ is the clipped PPO loss using the block-weighted advantages from Eq.~\ref{eq:tokenadv}, $\beta\mathcal{L}_{\rm KL}$ keeps the policy near the reference model, $\lambda_f\mathcal{L}_{\chi^2}$ is a chi-square regularizer that damps large positive log-ratio deviations, $\mathcal{L}_{\rm NLL}$ is the self-distillation loss on solved rollouts, and $c_v\mathcal{L}_V$ trains the block critic. The joint KL\,+\,$\chi^2$ design is motivated by the observation that KL regularization alone is too weak to prevent over-optimization of sparse rewards~\citep{huang2025chipomythos}. Appendix~\ref{app:blockcritic} gives the value-head features and calibration, Appendix~\ref{app:stability} defines $\mathcal{L}_{\rm NLL}$ and its schedule, and Appendix~\ref{app:trust} defines $\mathcal{L}_{\chi^2}$. Algorithm~\ref{alg:capr} summarizes one optimization iteration.

\paragraph{Inference.} At deployment, CAPR uses only the trained LoRA policy with the standard single-path dLLM denoising loop. Cache \& Steer, branching, pruning, the block critic, and self-distillation are training-only mechanisms. Their signal has already been distilled into the policy weights at train time. Inference therefore uses the same number of model forwards as the base dLLM, with no value head or training-time path state.


\section{Experiments}
\label{sec:exp}

We evaluate CAPR on four verifiable reasoning benchmarks under matched decoding budgets. The experiments ask three questions: whether block-level attribution improves final accuracy, whether the gains hold across dense and mixture-of-experts dLLMs, and whether the extra training signal is cheaper than tree-structured rollouts.

\subsection{Experimental Setup}
\label{sec:exp:setup}

\paragraph{Models and tasks.} We evaluate LLaDA-8B-Instruct~\citep{nie2025llada} and LLaDA-MoE-7B-A1B-Instruct~\citep{zhu2025lladamoe}. Both are trained with rank-$64$ LoRA adapters on frozen base weights, bf16 activations, fp32 optimizer state, and DeepSpeed ZeRO-2. Benchmarks are $4{\times}4$ Sudoku, Countdown, GSM8K~\citep{cobbe2021gsm8k}, and Math500~\citep{lightman2023lets}. Sudoku and Countdown stress exact constraint satisfaction, while GSM8K and Math500 test mathematical reasoning from natural-language prompts.

\paragraph{Evaluation.} We follow the \emph{d}-TreeRPO evaluation script~\citep{pan2025dtreerpo}: zero-shot pass@1, deterministic decoding ($\tau{=}0$), block length $\ell_{\rm blk}{=}32$, and denoising steps $N{=}L/2$. 


\newcommand{\caprscore}[2]{\shortstack{\textbf{#1}\\\textcolor{gray}{\scriptsize +#2}}}
\begin{table}[t]
\centering
\scriptsize
\caption{Pass@1 on four reasoning benchmarks at $256$- and $512$-token generation budgets. CAPR is evaluated with the same zero-shot setting, including prompts, parser, deterministic decoding, and token budget as in \emph{d}-TreeRPO. Gray values show gains over the corresponding base model.}
\label{tab:main}
\tabcolsep3pt
\begin{tabular}{@{\extracolsep{\fill}}l cc cc cc cc@{}}
\toprule
\textbf{Method} & \multicolumn{2}{c}{\textbf{Sudoku}} & \multicolumn{2}{c}{\textbf{Countdown}} & \multicolumn{2}{c}{\textbf{GSM8K}} & \multicolumn{2}{c}{\textbf{Math500}} \\
\cmidrule(lr){2-3}\cmidrule(lr){4-5}\cmidrule(lr){6-7}\cmidrule(lr){8-9}
 & \textbf{256} & \textbf{512} & \textbf{256} & \textbf{512} & \textbf{256} & \textbf{512} & \textbf{256} & \textbf{512} \\
\midrule
\multicolumn{9}{@{}l}{\textit{LLaDA-8B-Instruct}} \\
\midrule
Base                       & 6.7 & 5.5 & 19.5 & 16.0 & 76.7 & 78.2 & 32.4 & 36.2 \\
\;+ Diffu-GRPO             & 12.9 & 11.2 & 31.3 & 37.1 & 79.8 & 81.9 & 34.1 & 39.0 \\
\;+ VRPO                   & 12.8 & 9.6 & 22.3 & 18.0 & 80.1 & 81.5 & 35.6 & 34.8 \\
\;+ wd1                    & 25.2 & 24.2 & 51.2 & 46.1 & 80.8 & 82.3 & 34.4 & 39.0 \\
\;+ SAPO        & 20.3 & 16.1 & 52.0 & 56.3 & 80.6 & 82.1 & 33.8 & 38.4 \\
\;+ d2-stepMerge & 76.1 & 66.2 & 52.4 & 52.1 & 81.1 & 82.0 & 34.4 & 38.5 \\
\;+ TraceRL                & 25.6 & 25.4 & 50.4 & 52.6 & 80.3 & 82.4 & 35.6 & 39.1 \\
\;+ \emph{d}-TreeRPO       & 92.9 & 80.3 & 71.1 & 62.1 & 81.2 & 82.6 & 37.7 & 38.9 \\
\rowcolor{gray!15}
\;+ \textbf{CAPR}
  & \caprscore{93.0}{86.3} & \caprscore{92.7}{87.2}
  & \caprscore{72.7}{53.2} & \caprscore{69.9}{53.9}
  & \caprscore{82.4}{5.7} & \caprscore{82.4}{4.2}
  & \caprscore{37.9}{5.5} & \caprscore{39.8}{3.6} \\
\midrule
\multicolumn{9}{@{}l}{\textit{LLaDA-MoE-7B-A1B-Instruct}} \\
\midrule
Base                       & 12.7 & 8.1 & 42.6 & 41.4 & 71.8 & 71.0 & 30.1 & 42.2 \\
\;+ Diffu-GRPO             & 17.3 & 15.4 & 50.1 & 48.2 & 74.2 & 73.7 & 38.1 & 43.4 \\
\;+ wd1                    & 40.6 & 35.4 & 56.6 & 58.7 & 76.2 & 76.4 & 39.8 & 43.4 \\
\;+ SAPO         & 21.2 & 18.7 & 54.2 & 54.8 & 75.2 & 76.8 & 38.6 & 40.4 \\
\;+ GDPO                   & 31.2 & 30.6 & 58.1 & 53.4 & 74.8 & 75.2 & 38.4 & 41.2 \\
\;+ TraceRL                & 27.4 & 25.2 & 54.2 & 49.1 & 75.4 & 76.1 & 40.0 & 44.1 \\
\;+ \emph{d}-TreeRPO       & 78.3 & 72.0 & 67.2 & 60.6 & 75.5 & 77.4 & 41.2 & 46.3 \\
\rowcolor{gray!15}
\;+ \textbf{CAPR}
  & \caprscore{84.8}{72.1} & \caprscore{81.0}{72.9}
  & \caprscore{68.8}{26.2} & \caprscore{72.7}{31.3}
  & \caprscore{79.4}{7.6} & \caprscore{77.5}{6.5}
  & \caprscore{41.6}{11.5} & \caprscore{50.4}{8.2} \\
\bottomrule
\end{tabular}
\end{table}

\paragraph{Baselines and hyperparameters.} Baselines include the base model, Diffu-GRPO~\citep{zhao2025d1}, VRPO~\citep{zhu2025llada1_5}, wd1~\citep{tang2025wd1}, SAPO~\citep{xie2025step}, d2-stepMerge~\citep{wang2025d2}, TraceRL~\citep{wang2025revolutionizing}, GDPO where available, and \emph{d}-TreeRPO~\citep{pan2025dtreerpo}. CAPR uses $G{=}8$ rollouts, Branch \& Prune with fork fraction $f_b{=}0.5$ and $G_b{=}2$ sibling continuations, GAE $(\gamma,\lambda_{\rm gae}){=}(0.99,0.95)$, value coefficient $c_v{=}0.5$, chi-square regularizer coefficient $\lambda_f{=}0.01$, reference-KL coefficient $\beta{=}0.01$, learning rate $3{\times}10^{-5}$, and max self-distillation weight $w_{\rm sd}^{\max}{=}0.10$. Baselines reproduced from \emph{d}-TreeRPO~\citep{pan2025dtreerpo}. App.~\ref{app:hyper} gives the full configuration.

\subsection{Main Results}
\label{sec:exp:main}

Table~\ref{tab:main} reports results for the two decoding budgets $L{=}256$ and $L{=}512$ in separate columns so short- and long-budget decoding are compared under the same parser, prompt format, and token limit for every method.
At $L{=}256$, CAPR is best or tied-best on all reported tasks for both backbones. 

On LLaDA-8B-Instruct, at $L{=}256$, CAPR reaches $93.0$ on Sudoku, $72.7$ on Countdown, $82.4$ on GSM8K, and $37.9$ on Math500. Compared with \emph{d}-TreeRPO at $L{=}256$, the gains are $+0.1$, $+1.6$, $+1.2$, and $+0.2$ points, respectively. The largest absolute improvements over flat-rollout baselines occur on Sudoku and Countdown, where the final reward is sparse and a single rollout-level advantage gives weak reward assignment.

\begin{figure}[!t]
    \centering
    \includegraphics[width=\linewidth]{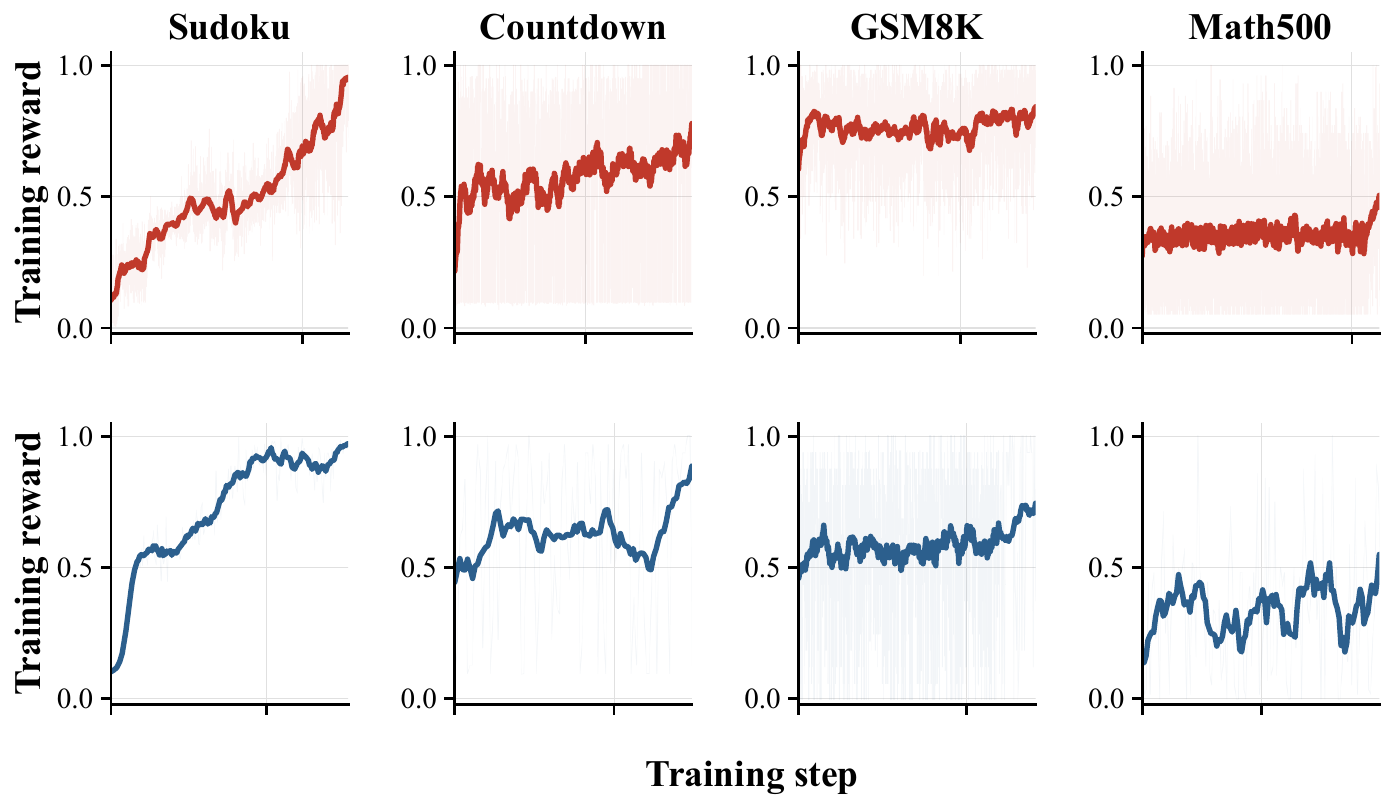}
    \caption{Training reward for the eight CAPR runs. Curves are centred rolling means with raw reward shown faintly. Sudoku shows the clearest cold start, GSM8K and Math500 start from stronger base-model competence. (Top) LLaDA-8B-Instruct (Bottom) LLaDA-MoE-7B-A1B-Instruct. Refer App.~\ref{app:reward-comparison} for contrast with baselines}
    \label{fig:maincurves}
\end{figure}

Fig.~\ref{fig:maincurves} shows that training rewards improve consistently across most tasks, with Sudoku exhibiting the strongest cold-start effect and the steepest gains during training. In contrast, GSM8K and Math500 begin from substantially higher initial rewards, indicating stronger base-model competence but comparatively smaller RL-driven improvements. The MoE backbone generally shows noisier optimization dynamics than the dense model, though both converge to stable high-reward regimes. 

On LLaDA-MoE-7B-A1B-Instruct, CAPR obtains $84.8$ on Sudoku, $68.8$ on Countdown, $79.4$ on GSM8K, and $41.6$ on Math500. This is $+6.5$ over \emph{d}-TreeRPO on Sudoku and $+1.6$ on Countdown, $+4.1$ on GSM8k while slightly improving the best prior scores on Math500. Thus, same training recipe transfers from dense backbone to sparse-MoE backbone without task-specific retuning. Case study in App.~\ref{app:case}

\begin{table}[t]
\centering
\scriptsize
\caption{Compute cost on $4{\times}4$ Sudoku with LLaDA-8B-Instruct. \textbf{Batch} = wall-time per rollout batch; \textbf{Update} = wall-time per parameter update; \textbf{Conv.} = total training time to convergence. Baselines reproduced from~\citep{pan2025dtreerpo}.  For all batch size = 4/device }
\label{tab:compute}
\begin{tabular}{lcccc}
\toprule
\textbf{Method} & \textbf{Batch (s)} & \textbf{Update (s)} & \textbf{Conv. (h)} & \textbf{Acc (\%)} \\
\midrule
Diffu-GRPO       & 111 & \phantom{0}9.12 & $\approx$24 & 12.9 \\
wd1              & \phantom{0}86 & \phantom{0}7.21 & $\approx$24 & 25.2 \\
SAPO             & 423 & 35.25           & $\approx$72 & 20.3 \\
GDPO             & 153 & 12.72           & $\approx$48 & 25.7 \\
TraceRL          & 604 & 43.14           & $\approx$48 & 25.6 \\
\emph{d}-TreeRPO & 596 & \phantom{0}9.91 & $\approx$48 & 92.9 \\
\rowcolor{gray!15}
\textbf{CAPR}    & \textbf{165} & \phantom{0}\textbf{10.10} & $\approx$\textbf{24} & \textbf{93.0} \\
\bottomrule
\end{tabular}
\end{table}
Table~\ref{tab:compute} reports wall-clock cost on $8{\times}$A100-80GB. CAPR takes $165$ seconds per optimization step versus $596$ seconds for \emph{d}-TreeRPO. The saving comes from sharing the early denoising prefix and spending extra forwards only on a small number of suffix continuations, while the block critic supplies the per-block attribution.
\begin{table}[!t]
\centering
\scriptsize
\caption{Module ablation on $4{\times}4$ Sudoku (LLaDA-8B-Instruct). Br/Pr/Cr/SD denote Branch, Prune, Block Critic, and self-distillation. Metrics are computed over the last $200$ steps.}
\label{tab:ablation}
\begin{tabular}{@{\extracolsep{\fill}}cccc cccc@{}}
\toprule
\textbf{Br} & \textbf{Pr} & \textbf{Cr} & \textbf{SD} & $r_{\rm peak}$ & $r_{\text{last200}}$ & $D_{\rm KL}$ & $\mathrm{zsr}$ \\
\midrule
\checkmark &            &            &            & 0.27 & 0.20 & 0.61 & 0.00 \\
\checkmark & \checkmark &            &            & 0.36 & 0.34 & 1.30 & 0.00 \\
           &            & \checkmark &            & 0.48 & 0.50 & 3.07 & 0.00 \\
\checkmark &            & \checkmark &            & 0.53 & 0.46 & 0.81 & 0.00 \\
\checkmark & \checkmark & \checkmark &            & 0.85 & 0.69 & 0.16 & 0.00 \\
\checkmark & \checkmark & \checkmark & \checkmark & 1.00 & 0.83 & 1.15 & 0.22 \\
\rowcolor{gray!15}
\checkmark & \checkmark & \checkmark & \checkmark & \textbf{0.90} & \textbf{0.83} & \textbf{0.21} & \textbf{0.10} \\
\bottomrule
\end{tabular}
\end{table}
\vspace{-0.3cm}
\subsection{Ablation: Contribution of Each Mechanism}
\label{sec:exp:abl}
\begin{figure}[!!t]
    \centering
    \includegraphics[width=\linewidth]{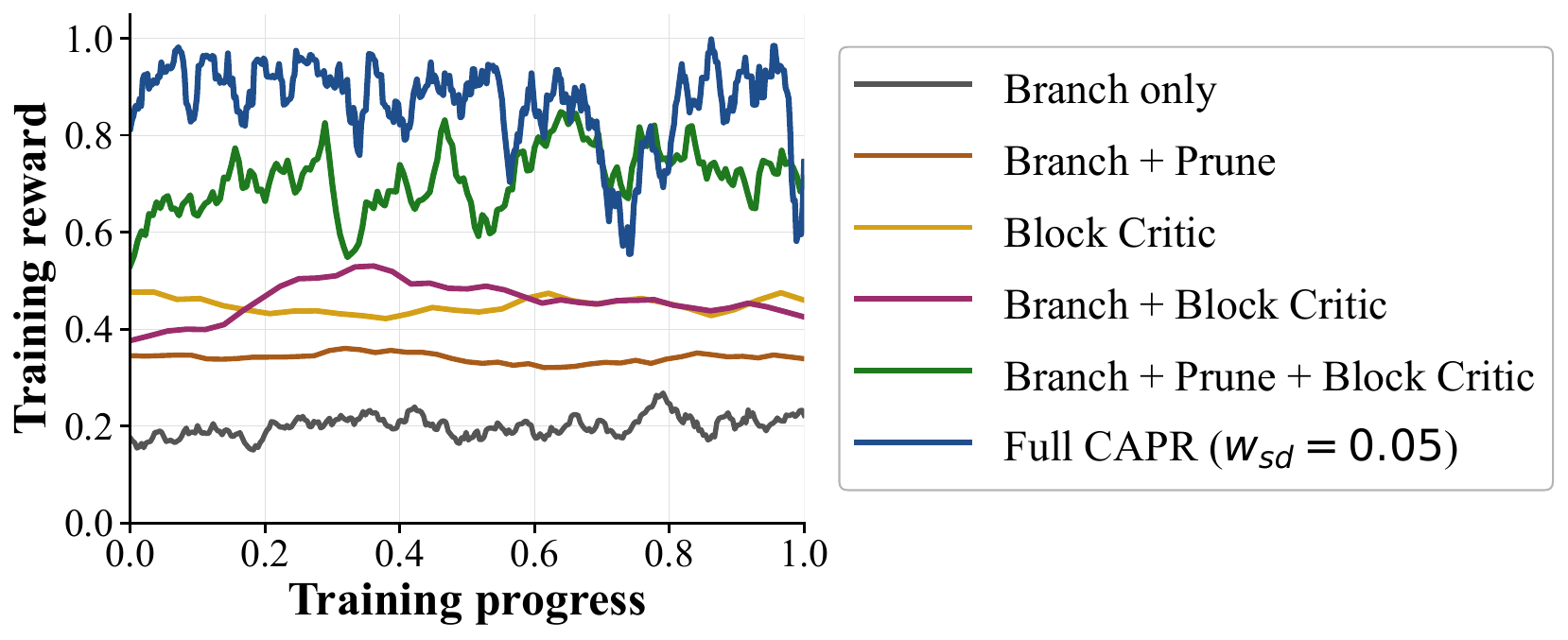}
    \vspace{-1.0em}
    \caption{Sudoku training reward under progressive activation of CAPR mechanisms. Branching and pruning alone give small gains, the block critic changes the slope, and self-distillation moves the curve into the high-reward band.}
    \label{fig:sudokuabl}
\end{figure}

\begin{figure}[!t]
    \centering
    \includegraphics[width=0.8\linewidth]{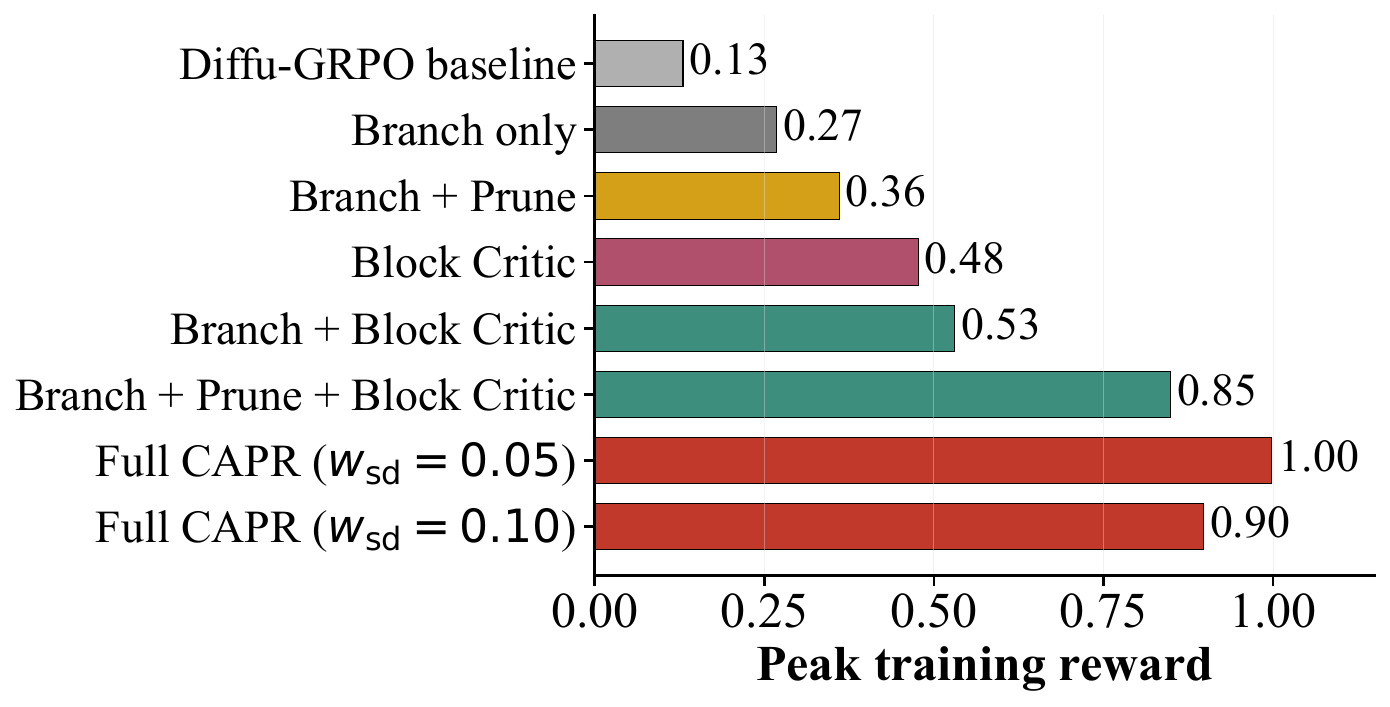}
    \caption{Peak Sudoku reward by ablation configuration. The bar chart mirrors Table~\ref{tab:ablation}. The largest jump occurs when Branch \& Prune is paired with the Block Critic, followed by self-distillation.}
    \label{fig:contrib}
\end{figure}

Table~\ref{tab:ablation} and Figures~\ref{fig:sudokuabl} and~\ref{fig:contrib} isolate the CAPR mechanisms on Sudoku, where sparse rewards make reward assignment most visible. Branch-only training improves over Diffu-GRPO but stays in the low-reward regime, adding pruning gives only a modest lift. The block critic is the first component that changes the learning curve substantially. By itself it reaches $0.48$ peak reward, and together with Branch \& Prune it reaches $0.85$. This shows that selective suffix exploration and per-block attribution are complementary rather than interchangeable.

The full model adds self-distillation on top of this branch--critic stack. In Figure~\ref{fig:sudokuabl}, the full CAPR curve rises into the high-reward band early and remains above the non-distilled variants for most of training. Figure~\ref{fig:contrib} shows the same progression as peak rewards: $0.27$ for Branch only, $0.36$ with pruning, $0.85$ with Branch \& Prune plus the Block Critic, and $0.90$--$1.00$ for full CAPR depending on maximum self-distillation weight $w_{\rm sd}^{\max}$. We select $w_{\rm sd}^{\max}{=}0.10$ because it keeps the last-$200$ reward equal to the highest-peak variant while reducing KL from $1.15$ to $0.21$.

\begin{figure}[!!t]
    \centering
    \includegraphics[width=\linewidth]{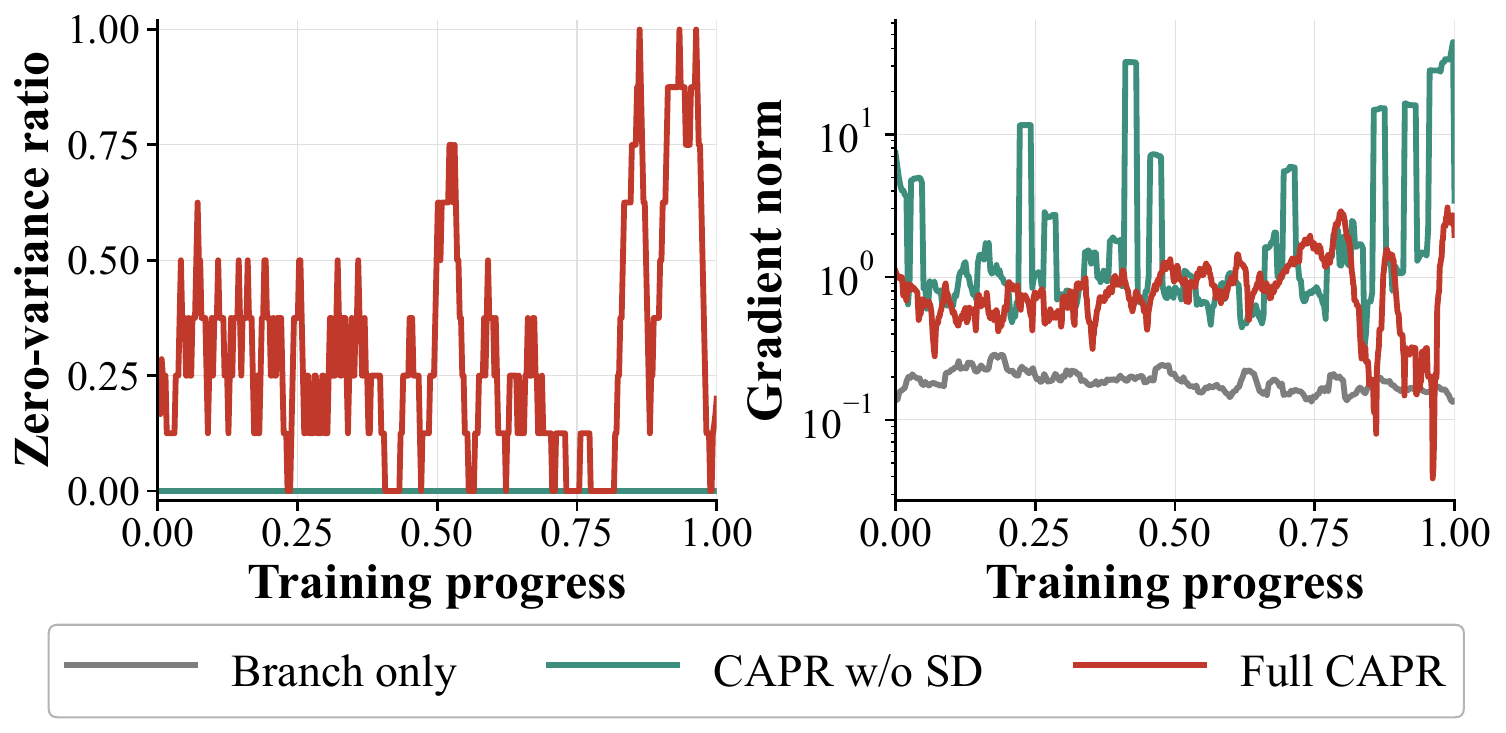}
    \caption{Sudoku stability diagnostics. Left: self-distillation (SD) turns solved zero-variance groups into usable signal. Right: full CAPR avoids the large gradient spikes seen without self-distillation.}
    \label{fig:stability}
\end{figure}

Figure~\ref{fig:stability} explains the role of self-distillation. Without self-distillation, zero-variance groups contribute no relative advantage and therefore provide little training signal. Full CAPR reuses those solved groups as supervised targets, so the zero-variance ratio rises when more rollouts are solved instead of becoming dead data. The gradient panel shows the trade-off. Branch only has small gradients but weak learning, CAPR without self-distillation has large spikes, and full CAPR keeps useful solved-group signal while producing a smoother late-training gradient profile.

\subsection{Training Dynamics}
\label{sec:exp:dynamics}
\begin{table}[!t]
\centering
\scriptsize
\caption{Training-time diagnostics for the eight main CAPR experiments. Reward statistics use the last $200$ steps up to the selected peak; the remaining columns are medians over the same window.}
\label{tab:dynamics}
\begin{tabular}{@{\extracolsep{\fill}}l cccccc@{}}
\toprule
\textbf{Benchmark} & $r_{\rm peak}$ & $r_{\text{last200}}$ & $D_{\rm KL}$ & $\chi^2_{\rm reg}$ & $\mathcal{L}_V{\times}10^3$ & $|\nabla|$ \\
\midrule
\multicolumn{7}{l}{\textit{LLaDA-8B-Instruct}} \\
Sudoku    & 0.90 & 0.83 & 0.29 & 0.25 & \phantom{0}1.0 & 0.00 \\
Countdown & 0.87 & 0.63 & 0.07 & 0.18 & \phantom{0}4.0 & 0.05 \\
GSM8K     & 0.81 & 0.78 & 0.06 & 0.14 & \phantom{0}6.0 & 0.08 \\
Math500   & 0.45 & 0.36 & 0.07 & 0.21 & \phantom{0}5.0 & 0.06 \\
\midrule
\multicolumn{7}{l}{\textit{LLaDA-MoE-7B-A1B-Instruct}} \\
Sudoku    & 0.97 & 0.72 & 1.02 & 0.09 & 17.0 & 0.13 \\
Countdown & 0.94 & 0.60 & 0.57 & 0.15 & 13.0 & 0.13 \\
GSM8K     & 0.86 & 0.64 & 0.06 & 0.28 & \phantom{0}7.0 & 0.12 \\
Math500   & 0.55 & 0.34 & 0.97 & 0.03 & \phantom{0}7.0 & 0.14 \\
\bottomrule
\end{tabular}
\end{table}

\begin{figure}[!!t]
    \centering
    \includegraphics[width=\linewidth]{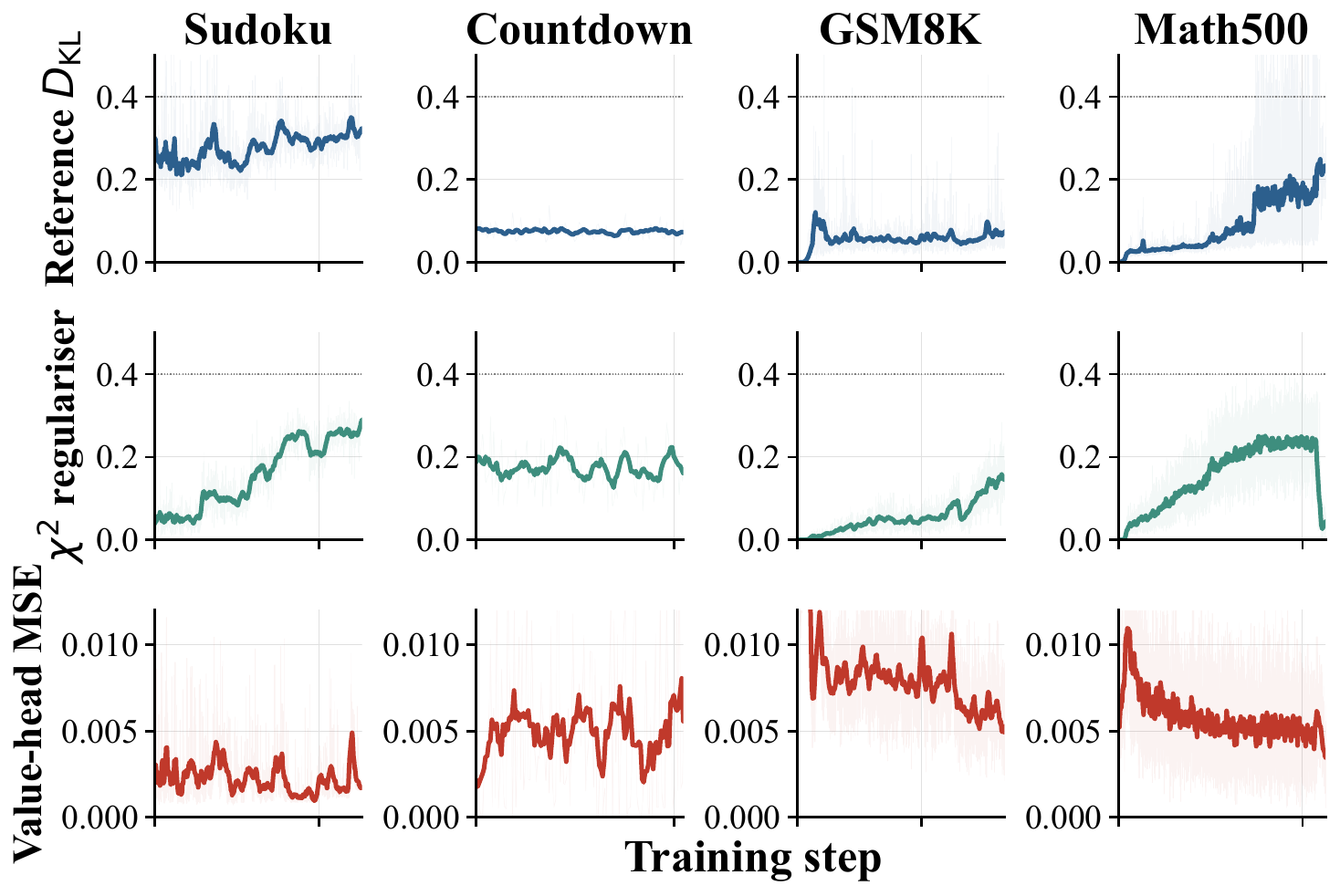}
    \caption{Dense-backbone diagnostics. Reference KL stays below the guide line, the chi-square regularizer remains bounded, and the value-head MSE stays small across benchmarks.}
    \label{fig:dynamics}
\end{figure}

Table~\ref{tab:dynamics} and Figure~\ref{fig:dynamics} check whether the accuracy gains come from unstable policy drift. For the dense backbone, reference KL stays below the $0.4$ guide line in all four tasks. Countdown and GSM8K remain especially close to the reference model, while Sudoku and Math500 move more because they require larger changes from the base policy. The chi-square regularizer statistic rises when learning accelerates but stays bounded, and the value-head MSE remains small, indicating that the block critic is tracking the changing policy rather than diverging.

\begin{figure}[!!t]
    \centering
    \includegraphics[width=\linewidth]{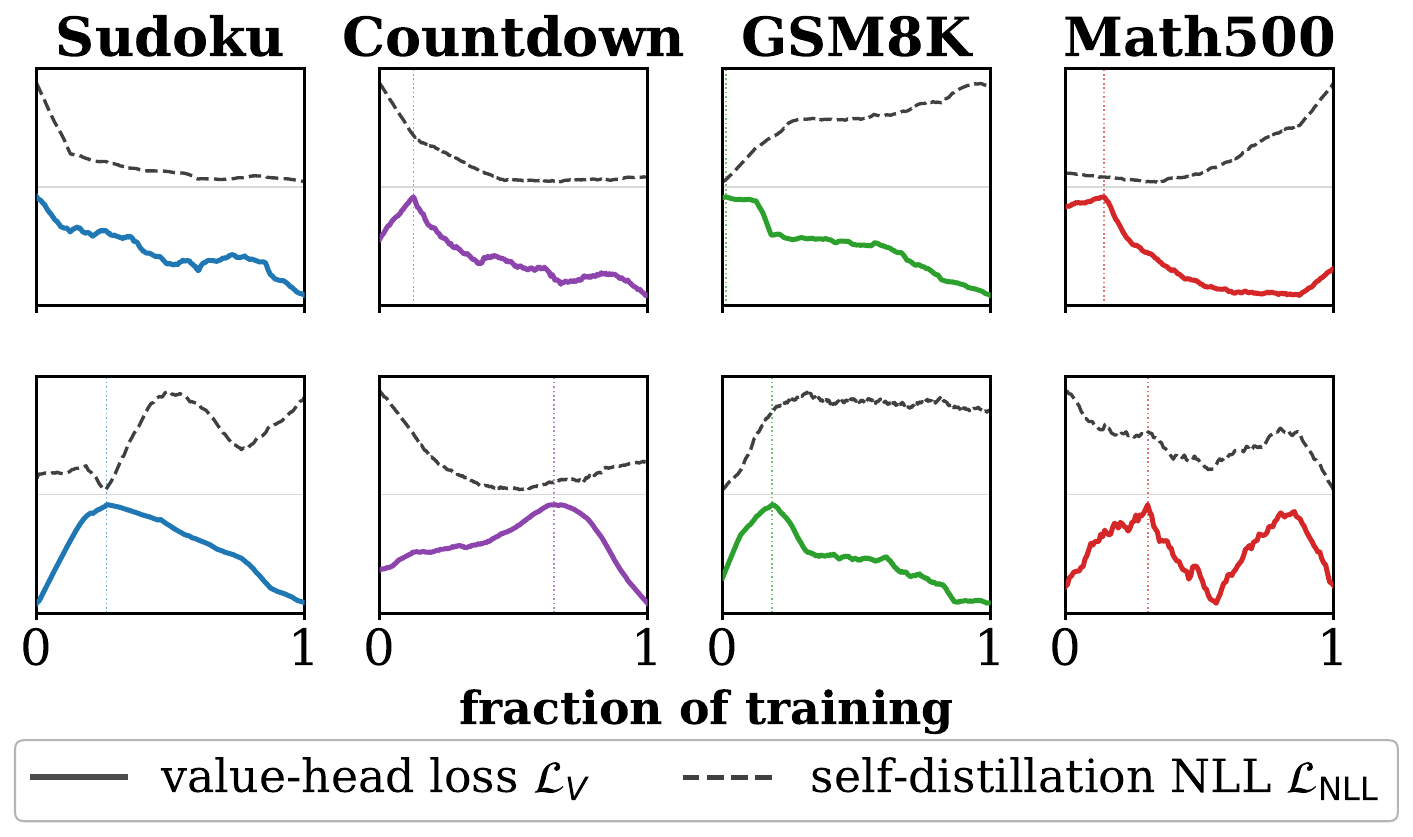}
    \caption{Block Critic loss and self-distillation NLL for the eight CAPR runs. Curves are min-max normalized within each panel, peaks mark phases where the policy changes fastest. (Top) LLaDA-8B-Instruct (Bottom) LLaDA-MoE-7B-A1B-Instruct.}
    \label{fig:vhead}
\end{figure}

The MoE rows in Table~\ref{tab:dynamics} show the same qualitative pattern with higher task-dependent KL on Sudoku and Math500, which is expected for routing-sensitive logits. Importantly, the reported checkpoints still have controlled gradient norms and stable last-window rewards. Figure~\ref{fig:vhead} gives the complementary view. The block critic loss typically peaks when the policy is changing fastest and then falls or stabilises, while the self-distillation NLL tracks how predictable solved rollouts become. These modules are therefore useful as training-time attribution and stabilization signals, but they are removed at inference, only the trained LoRA policy is used.

\subsection{Hyperparameter Sensitivity}
\label{sec:exp:sens}
\begin{table}[!t]
\centering
\scriptsize
\caption{Hyperparameter sensitivity on $4{\times}4$ Sudoku (LLaDA-8B-Instruct). Entries are peak training reward for maximum self-distillation weight $w_{\rm sd}^{\max}$ and learning rate $\eta$, other settings are in App.~\ref{app:hyper}.}
\label{tab:sensitivity}
\begin{tabular}{l cc}
\toprule
\textbf{Configuration} & $\eta{=}3{\times}10^{-5}$ & $\eta{=}5{\times}10^{-5}$ \\
\midrule
$w_{\rm sd}^{\max}{=}0$        & 0.85  & 0.63 \\
$w_{\rm sd}^{\max}{=}0.05$     & 1.00  & 0.85 \\
\rowcolor{gray!15}
$w_{\rm sd}^{\max}{=}0.10$     & \textbf{0.90}  & 0.78 \\
\bottomrule
\end{tabular}%
\end{table}
Table~\ref{tab:sensitivity} varies the two most important optimization knobs on Sudoku. The lower learning rate is consistently better. The selected setting, $w_{\rm sd}^{\max}{=}0.10$ and $\eta{=}3{\times}10^{-5}$, gives slightly lower peak reward than $w_{\rm sd}^{\max}{=}0.05$ but substantially tighter KL and smoother gradients, hence used in all exps.

\section{Conclusion}
\label{sec:conclusion}
Masked diffusion language models expose a denoising trace that standard RL post-training largely ignores. CAPR turns this trace into a training signal. It records confidence, entropy, and token stability in a compact path state, uses that state to steer reverse denoising steps, reuses cached prefixes to create cheap sibling continuations, and trains a Block Critic to convert one final outcome reward into per-block PPO weights. In this way, CAPR preserves the locality of tree-style comparisons without expanding a full search tree.

Across dense and mixture-of-experts LLaDA backbones, CAPR improves reasoning under the same decoding setting as the baselines. At the matched 256-token budget, it is best or tied-best on all reported tasks, with the largest gains on Sudoku and Countdown, where flat rollout-level signals are weakest. The 512-token results show that the same pattern holds at longer generation lengths. On Sudoku, CAPR matches the strongest tree-structured baseline while using less than one third of its per-step wall-clock time. The ablations show that the gains do not come from branching alone. Branching and pruning add useful diversity, the Block Critic provides the largest intermediate improvement by localizing the reward signal, self-distillation keeps solved groups useful after group-relative advantages vanish, and the chi-square regularizer limits unstable ratio spikes. CAPR therefore offers a practical middle ground for dLLM-RL, with finer reward assignment than flat rollouts at substantially lower cost than full tree expansion.

\section*{Limitations}

CAPR is evaluated on automatically verifiable reasoning tasks. This is a controlled setting for studying reward assignment, but it does not establish performance for open-ended instruction following, factuality, dialogue safety, or preference-based objectives where rewards are noisier and less local. The experiments also use zero-shot pass@1 decoding, pass@$k$ evaluation, robustness under different sampling settings, and out-of-domain prompts remain important tests of generalization.

The block critic is learned online from the same rollouts used to update the policy. This keeps CAPR lightweight, but the value estimates may be less reliable for very long reasoning chains, delayed subgoals, or domains where many partial trajectories can lead to the same final reward. CAPR is also tested only on LLaDA-family masked diffusion language models. Other dLLM architectures may expose different trace statistics, and branch fraction, block size, self-distillation weight, and value-head calibration may need retuning.

Finally, the compute measurements reflect our software stack and $8{\times}$A100-80GB hardware. CAPR reduces denoising continuations relative to tree expansion, but it still requires RL fine-tuning of large language models. Memory layout, kernel efficiency, and distributed-training overhead can change wall-clock cost on other systems.

\section*{Ethical Considerations}

CAPR is a training algorithm for improving reasoning in masked diffusion language models. The experiments use public reasoning benchmarks with automatically checkable rewards, and we do not introduce new user data or sensitive annotations. However, stronger reasoning models can be misused in the same ways as other capable language models, including producing misleading explanations, assisting harmful workflows, or making incorrect answers appear more persuasive. CAPR does not add a safety-alignment layer, so user-facing deployments should retain the safeguards, monitoring, refusal behaviour, and domain-specific evaluations required for the underlying model.

RL fine-tuning large models also has an environmental cost. CAPR is designed to reduce training cost relative to tree-structured rollouts, and we report hardware and wall-clock measurements to make that cost visible. Future work should continue to report compute transparently and study whether trajectory-aware signals can be reused across tasks, seeds, or training runs.

We downloaded models from huggingface. GSAI-ML/LLaDA-8B-Instruct\footnote{\url{https://huggingface.co/GSAI-ML/LLaDA-8B-Instruct}} is available under MIT license. 
inclusionAI/LLaDA-MoE-7B-A1B-Instruct\footnote{\url{https://huggingface.co/inclusionAI/LLaDA-MoE-7B-A1B-Instruct}} is available under Apache license 2.0. The models were used as per their intended use.

\bibliography{custom}

\newpage
\appendix

\input{tex/appendix-algorithm}
\input{tex/appendix-complexity}
\input{tex/appendix-hyper-capr}
\input{tex/appendix-proof}
\input{tex/appendix-baselines}
\input{tex/appendix-reward-capr}

\input{tex/appendix-prompt}
\input{tex/appendix-case}

\end{document}

%% file: tex/appendix-algorithm.tex
\section{CAPR Optimisation Iteration}
\label{app:algorithm}

Algorithm~\ref{alg:capr} states the full per-prompt update used in our experiments. The same loop runs unchanged across the four benchmarks. Only the schedules and weights listed in Table~\ref{tab:hyper} differ. Symbols follow \S\ref{sec:method}: $G$ is the target group size, $f_b$ the branch fraction, $G_b$ the number of branches per cached prefix, $B$ the number of blocks, $\tau_p$ the Polyak rate of the slow value head $V_{\bar\phi}$, $\beta_{\rm TD}$ the TD blend, and $\tau_{\rm sd}$ the self-distillation reward threshold.

\subsection{Path-State Update Details}
\label{app:metastate}

For masked position $j$ at denoising state $z_t$, let $p_j^{(t)}=\pi_\theta(\cdot\mid z_t)_j$ and $v_j^{(t)}=\arg\max_v p_j^{(t)}(v)$. Let $\bar v_j^{(t-1)}$ be the token prediction carried from the previous denoising update. CAPR forms the same path-state features used in \S\ref{sec:controller}:
\begin{equation}
\begin{gathered}
c_j^{(t)}=p_j^{(t)}(v_j^{(t)}),\qquad
s_j^{(t)}=\indic[v_j^{(t)}=\bar v_j^{(t-1)}],\\
e_j^{(t)}=\frac{H(p_j^{(t)})}{\max_k H(p_k^{(t)})},
\qquad
a_j^{(t)}=c_j^{(t)}(1-e_j^{(t)}).
\end{gathered}
\label{eq:metastate_features}
\end{equation}
Here $c_j^{(t)}$ is confidence, $s_j^{(t)}$ is stability, $e_j^{(t)}$ is normalized entropy, and $a_j^{(t)}$ is certainty. The PathGRU update in Eq.~\ref{eq:pathgru} is implemented as
{\small
\begin{align}
r_j^{(t)} &= \sigma\!\left(\alpha_e(0.5-e_j^{(t)})\right), \nonumber\\[-0.2em]
u_j^{(t)} &= \sigma\!\left(\alpha_s\left(\frac{s_j^{(t)}+a_j^{(t)}}{2}-0.5\right)\right), \nonumber\\[-0.2em]
m_j^{(t)} &= \tfrac14(2c_j^{(t)}-1)(1+a_j^{(t)})(1+s_j^{(t)}), \nonumber\\[-0.2em]
\tilde h_j^{(t)} &= \tanh\!\left(r_j^{(t)} h_j^{(t-1)} + (1-r_j^{(t)})m_j^{(t)}\right), \nonumber\\[-0.2em]
h_j^{(t)} &= \operatorname{RMSNorm}\!\left(u_j^{(t)}h_j^{(t-1)}+(1-u_j^{(t)})\tilde h_j^{(t)}\right),
\label{eq:metastate_update}
\end{align}
}
where $r_j^{(t)}$ is the reset gate and $u_j^{(t)}$ is the update gate, $\alpha_e$ and $\alpha_s$ are constants. The final RMS normalization is applied across positions in the sequence and prevents prompt-dependent hidden-scale drift.

\subsection{Block Critic Features and TD Blending}
\label{app:blockcritic}

We use $b\in\{1,\ldots,B\}$ for block index and reserve $t$ for denoising time. At block boundary $b$, the value head receives a compact state with both scalar features and a pooled token-embedding feature:
\begin{equation}
\small
\psi_b =
\big(
\mu(h_b),\sigma(h_b),
\rho^{\rm mask}_b,
b/(B{-}1),
\bar R_q,\sigma_R(q),
e_b
\big),
\end{equation}
where $\mu(h_b)$ and $\sigma(h_b)$ are the mean and standard deviation of the path state inside the current block, $\rho^{\rm mask}_b$ is the fraction of positions in that block that remain masked, $(\bar R_q,\sigma_R(q))$ are the mean and standard deviation of outcome rewards among rollouts for prompt $q$, and $e_b$ is the average input embedding of the tokens that have already been placed in block $b$.

For rollout $i$ with final outcome reward $R_i$, let $n_{i,b}$ be the number of tokens unmasked in block $b$, i.e., positions that switch from \texttt{[MASK]} to a visible token while block $b$ is being denoised. CAPR forms reward-conserving block rewards
\begin{equation}
r_{i,b}
=
\omega_{\rm blk} R_i
\frac{n_{i,b}}{\sum_{c=1}^{B} n_{i,c}}
+
(1-\omega_{\rm blk})R_i\indic[b=B].
\label{eq:block_reward_appendix}
\end{equation}
The first term gives dense block supervision proportional to the tokens revealed in each block, while the second term keeps the terminal outcome signal on the final block. This mattered in runs that reach EOS before block $B$, instead of $1/B$ it will be accurate with $n_{i,b}/\sum_c n_{i,c}$, where $n_{i,b}$ is the number of tokens latest unmasked in block $b$.

The target head $V_{\bar\phi}$ computes TD residuals and GAE advantages on the block-boundary chain~\citep{schulman2015high,sutton1988learning}:
\begin{align}
\delta_{i,b}
&=
r_{i,b}
+\gamma V_{\bar\phi}(s_{i,b+1})
-V_{\bar\phi}(s_{i,b}), \nonumber\\[-0.2em]
\hat A^\lambda_{i,b}
&=
\sum_{\ell=0}^{B-b}
(\gamma\lambda_{\rm gae})^\ell
\delta_{i,b+\ell},
\label{eq:block_gae_appendix}
\end{align}
with $V_{\bar\phi}(s_{i,B+1})=0$. The value-head regression target is the corresponding $\lambda$-return:
\begin{equation}
y^V_{i,b}
=
\operatorname{sg}
\left(
\hat A^\lambda_{i,b}
+
V_{\bar\phi}(s_{i,b})
\right),
\label{eq:block_value_target_appendix}
\end{equation}
where $\operatorname{sg}$ denotes stop-gradient. The online value head $V_\phi$ is trained to predict $y^V_{i,b}$, while $V_{\bar\phi}$ is updated as a slow Polyak average~\citep{lillicrap2015continuous}.

The block-level signal is blended into PPO only when the value head is calibrated. Let $\overline{\mathcal{L}}_V$ be the recent value-head loss and $b_V$ be a running baseline tracking the best smoothed value loss so far. CAPR uses
\begin{equation}
q_V =
\frac{b_V^2}{b_V^2+\overline{\mathcal{L}}_V^2},
\qquad
\beta_{\rm TD}^{\rm eff}
=
\beta_{\rm TD}\, q_V.
\label{eq:tdgate}
\end{equation}
When the value head loss rises far above its baseline, $q_V$ becomes small and CAPR falls back toward the flat GRPO signal. As the value head recovers, the block weighting reopens automatically.

\subsection{Self-Distillation Loss}
\label{app:stability}

CAPR uses self-distillation only during training. Let $\mathcal{A}_i$ be the active completion tokens of rollout $i$, i.e., the valid completion tokens selected by the current PPO mask. In our reported runs all tokens in $\mathcal{A}_i$ are averaged uniformly, so no additional token weighting is used.

First, solved rollouts supervise themselves with a small next-token loss. Let $\mathcal{C}=\{i:R_i\ge\tau_{\rm sd}\}$ be the solved rollouts and let $\rho_{\rm cor}=|\mathcal{C}|/(QG)$ be their fraction in a batch of $Q$ prompts. CAPR uses
\begin{equation}
\mathcal{L}_{\rm NLL}
= -w_{\rm sd}^{\rm eff}
\frac{\sum_{i\in\mathcal{C}}\sum_{j\in\mathcal{A}_i}\log\pi_\theta(y_{ij})}
{\sum_{i\in\mathcal{C}}|\mathcal{A}_i|}.
\label{eq:nll}
\end{equation}
The effective self-distillation weight grows with the solved-rollout fraction and is bounded by the tuned maximum,
\begin{equation}
w_{\rm sd}^{\rm eff}
=\min\!\left(w_{\rm sd}\rho_{\rm cor}^{p},\quad w_{\rm sd}^{\max}\right),
\label{eq:sdweight}
\end{equation}
so it is near zero when solved rollouts are rare, increases as correct samples become common, and never exceeds $w_{\rm sd}^{\max}$. In Table~\ref{tab:sensitivity}, we vary this maximum while keeping the remaining self-distillation settings fixed.

\subsection{Chi-Square Regulariser}
\label{app:trust}

Sparse-reward updates can fail because of a few tokens, not because the whole completion moves too far. In a solved rollout, PPO may strongly increase the probability of a small number of already-likely tokens. These upper-tail log-ratio spikes can dominate the gradient, especially when most other rollouts have zero or identical reward. Standard PPO clipping and KL regularisation~\citep{kakade2002approximately,schulman2017proximal} control average policy movement, but they do not specifically target this token-level upper tail.

CAPR therefore adds a small outlier-only $\chi^2$ term. It is not a second full KL penalty: ordinary movement is still handled by PPO and KL. The extra term activates only when a token's centred log-ratio is unusually large and positive. This follows the robustness motivation of chi-squared preference optimisation~\citep{huang2025chipomythos} and is related in spirit to asymmetric policy-gradient controls that treat positive and negative update directions differently~\citep{yu2025dapo,xie2025step}. For each sequence, CAPR computes $d_j=\log\pi_\theta(y_j)-\log\pi_{\rm ref}(y_j)$ on the active completion-token set $\mathcal{A}$ and normalises it by
\begin{align}
\bar d &= |\mathcal{A}|^{-1}\sum_{k\in\mathcal{A}} d_k, \nonumber\\[-0.2em]
s_d &= \left(|\mathcal{A}|^{-1}\sum_{k\in\mathcal{A}}(d_k-\bar d)^2\right)^{1/2}, \nonumber\\[-0.2em]
\hat d_j &= (d_j-\bar d)/(s_d+\epsilon).
\end{align}
Only the upper tail is penalized:
\begin{equation}
\mathcal{L}_{\chi^2}
= |\mathcal{A}|^{-1}\sum_{j\in\mathcal{A}}\left[\hat d_j-1\right]_+^2.
\label{eq:chi2}
\end{equation}
The one-sided form is important. Positive spikes make the new policy over-commit to a few sampled tokens. Negative movements usually reflect probability mass being moved away from sampled tokens and are already bounded by the clipped PPO objective. Penalising both sides equally would make the regulariser behave like another broad KL term and would unnecessarily resist useful exploration.

A simple trust-region view gives the same intuition. For bounded token advantages, the performance-difference bound contains an error term proportional to how far the new policy moves from the old one~\citep{kakade2002approximately}. Since total variation is controlled by chi-squared divergence, keeping the upper-tail $\chi^2$ small bounds the harmful part of the update:
\begin{equation}
J(\pi_{\theta'}) - J(\pi_\theta)
\gtrsim
\widehat{\Delta}_{\rm PPO}
- C\sqrt{\mathcal{L}_{\chi^2}},
\end{equation}
for a constant $C$ depending on the advantage scale. CAPR uses Eq.~\ref{eq:chi2} only as a soft penalty, so the method keeps the usual PPO/KL update while damping the rare token-level jumps most likely to destabilise sparse-reward training.

\begin{algorithm}[h]
\caption{One CAPR optimisation iteration for prompt $q$.}
\label{alg:capr}
\small
\begin{algorithmic}[1]
\Require Policy $\pi_\theta$, reference $\pi_{\rm ref}$, value head $V_\phi$ and target head $V_{\bar\phi}$.
\Require Rollouts $G$, branches per cached prefix $G_b$, fork fraction $f_b$, denoising steps $N$, blocks $B$.
\State Set $M=\lceil G/G_b\rceil$ cached prefixes and initialise path state $h=\mathbf{0}$.
\For{$m=1,\ldots,M$}
    \State Denoise from $z_N$ to the fork step $f_bN$, updating $h$ from confidence, entropy, and stability.
    \State Cache the fork-step state and spawn $G_b$ sibling continuations by resampling the cached scores.
    \State Continue the siblings to completion with the same trace update and steering rule.
    \State At scheduled block boundaries, prune low-quality siblings using mean path-state quality.
\EndFor
\State Score the completed rollouts with the verifier to obtain outcome rewards $R_i$.
\For{each rollout $i$ and block $b$}
    \State Form the reward-conserving block signal
    \Statex \hspace{1.3em}$r_{i,b}=\omega_{\rm blk} R_i\frac{n_{i,b}}{\sum_c n_{i,c}}+(1-\omega_{\rm blk})R_i\indic[b=B]$.
    \State Compute TD residuals, GAE advantages $\hat A_{i,b}$, and value targets $y^V_{i,b}$ using Eq.~\ref{eq:block_gae_appendix} and Eq.~\ref{eq:block_value_target_appendix}.
\EndFor
\State Convert block advantages into token weights with Eq.~\ref{eq:tokenadv}.
\State Optimise the policy and value head with $\mathcal{L}_{\rm CAPR}$ in Eq.~\ref{eq:total}, including KL, $\chi^2$, and self-distillation terms.
\State Update the target head by Polyak averaging: $\bar\phi\leftarrow(1-\tau_p)\bar\phi+\tau_p\phi$.
\end{algorithmic}
\end{algorithm}

%% file: tex/appendix-complexity.tex
\section{Denoising Forward-Pass Cost}
\label{app:complexity}

We compare rollout-generation cost by counting denoising model forward passes for one prompt. This is the shared cost paid before the PPO-style update. For shared quantities we follow the \emph{d}-TreeRPO comparison setting~\citep{pan2025dtreerpo}: completion length $L{=}256$, and denoising budget $N{=}256$. CAPR-specific quantities use our reported values in Appendix~\ref{app:hyper}: branch fraction $f_b{=}0.5$ and branches per cached prefix $G_b{=}2$, $G{=}8$ terminal completions.

Flat methods sample each completion independently. With $G$ completions and $N$ denoising steps per completion, the cost is
\begin{equation}
C_{\rm flat}=G N.
\end{equation}
Thus the flat baseline costs $8\cdot256=2048$ forwards.

For \emph{d}-TreeRPO, we use the forward-count formula reported in \citet{pan2025dtreerpo}. With tree branching factor $B_{\rm tree}$ and depth $H$, the reported cost is
\begin{equation}
C_{\rm tree}
=\frac{B_{\rm tree}(B_{\rm tree}^{H}-1)}{B_{\rm tree}-1}\cdot\frac{N}{H}.
\end{equation}
With $B_{\rm tree}{=}4$, $H{=}2$, (these settings are the minimal requirements as per \citep{pan2025dtreerpo}) and $N{=}256$, this gives $(4+16)\cdot128=2560$ forwards.

CAPR uses a smaller comparison set by sharing the early prefix. One branch group runs a shared prefix for $f_bN$ steps, then forks $G_b$ sibling suffixes of length $(1-f_b)N$. Without pruning, one group costs
\begin{equation}
C_{\rm group}=\big[f_b+G_b(1-f_b)\big]N.
\end{equation}
Producing $G$ completions requires $G/G_b$ groups, giving
\begin{equation}
C_{\rm CAPR}
=\frac{G}{G_b}\big[f_b+G_b(1-f_b)\big]N.
\end{equation}
With $G{=}8$, $G_b{=}2$, $f_b{=}0.5$, and $N{=}256$, the no-pruning count is $4\cdot1.5\cdot256=1536$ forwards. Scheduled pruning can only reduce this value. If $0=p_0<p_1<\cdots<p_P<p_{P+1}=1$ are suffix checkpoints and $a_m$ siblings are kept on interval $[p_m,p_{m+1}]$, then one group costs
\begin{equation}
N\left[f_b+(1-f_b)\sum_{m=0}^{P}a_m(p_{m+1}-p_m)\right],
\end{equation}
which is no larger than the no-pruning value whenever $a_m\leq G_b$. Table~\ref{tab:complexity} therefore reports CAPR's conservative upper bound.

\begin{table}[h]
\centering
\small
\caption{Per-prompt denoising forward-pass cost under the \emph{d}-TreeRPO comparison setting. Shared quantities use, $L{=}256$, and $N{=}256$. CAPR-specific quantities use $G_b{=}2$ and $f_b{=}0.5$, $G{=}8$.}
\label{tab:complexity}
\setlength{\tabcolsep}{3.5pt}
\begin{tabular}{lcc}
\toprule
\textbf{Method} & \textbf{Forwards per prompt} & \textbf{\#\,fwd} \\
\midrule
Diffu-GRPO         & $G\,N$                                       & 2048 \\
\emph{d}-TreeRPO   & $\tfrac{B_{\rm tree}(B_{\rm tree}^{H}-1)}{B_{\rm tree}-1}\!\cdot\!\tfrac{N}{H}$ & 2560 \\
\textbf{CAPR}      & $\leq\tfrac{G}{G_b}\!\cdot\!\big[f_b{+}G_b(1{-}f_b)\big]\,N$ & $\leq\textbf{1536}$ \\
\bottomrule
\end{tabular}
\end{table}

%% file: tex/appendix-hyper-capr.tex
\section{Hyperparameters}
\label{app:hyper}

Table~\ref{tab:hyper} reports the CAPR hyperparameters used for the results in Table~\ref{tab:main}. Unless noted otherwise, the values are shared across tasks.

\begin{table}[h]
\centering
\caption{CAPR hyperparameters, organised by mechanism.}
\label{tab:hyper}
\setlength{\tabcolsep}{4pt}
\resizebox{\linewidth}{!}{%
\begin{tabular}{l l l}
\toprule
\textbf{Group} & \textbf{Parameter} & \textbf{Value} \\
\midrule
\multicolumn{3}{l}{\textit{Generation \& rollouts}} \\
                & target rollouts $G$              & 8 \\
                & block length $\ell_{\rm blk}$    & 32 tokens \\
                & number of blocks $B$             & $L/\ell_{\rm blk}$ \\
                & denoising steps $N$              & $L/2$ \\
                & evaluation temperature           & 0 / pass@1 decoding \\
                & rollout temperature              & 1.0 \\
\midrule
\multicolumn{3}{l}{\textit{Cache \& Steer path state (\S\ref{sec:controller})}} \\
                & $\alpha_e$ (entropy gate scale)  & 4.0 \\
                & $\alpha_s$ (stability gate scale)& 5.0 \\
                & $\gamma$ (path-state injection)  & 0.15 \\
                & $\lambda$ (path-state steering scale) & 1.0 \\
                & $\lambda_f$ ($\chi^2$ coefficient) & 0.01 \\
                & $\beta$ (reference KL coefficient) & 0.01 \\
\midrule
\multicolumn{3}{l}{\textit{Branch-and-prune sampler (\S\ref{sec:branching})}} \\
                & branch fraction $f_b$            & 0.5 \\
                & branches per cached prefix $G_b$ & 2 \\
                & first prune checkpoint $p_1$     & 0.33 \\
                & second prune checkpoint $p_2$    & 0.67 \\
                & keep ratio                       & 0.75 \\
\midrule
\multicolumn{3}{l}{\textit{Block Critic (\S\ref{sec:blockcritic})}} \\
                & TD discount $\gamma$             & 0.99 \\
                & GAE $\lambda_{\rm gae}$          & 0.95 \\
                & value-MSE coefficient $c_v$      & 0.5 \\
                & value-head hidden dim            & 128 \\
                & Polyak coefficient $\tau_p$      & 0.005 \\
                & replay-buffer size               & 32 \\
                & block-shaping mass $\omega_{\rm blk}$ & 0.20 \\
\midrule
\multicolumn{3}{l}{\textit{Self-distillation loss (Appendix~\ref{app:stability})}} \\
                & $w_{\rm sd}$ (base NLL scale)     & 0.10 \\
                & $w_{\rm sd}^{\max}$ (maximum NLL weight) & 0.10 \\
                & $\tau_{\rm sd}$ (reward threshold) & 0.99 \\
                & power $p$                        & 2.0 \\
                & $\tau_{\rm NLL}$ (NLL clamp)     & 3.0 \\
\midrule
\multicolumn{3}{l}{\textit{Optimisation}} \\
                & learning rate $\eta$              & $3{\times}10^{-5}$ \\
                & optimiser                         & AdamW \\
                & weight decay                      & 0.1 \\
                & gradient clip                     & 1.0 \\
                & PPO inner epochs $\mu$            & 2 \\
                & PPO clip $\epsilon$               & 0.4 \\
                & LoRA rank ($r$, $\alpha$, dropout)& $(128, 64, 0.05)$ \\
                & DeepSpeed stage                   & ZeRO-2 \\
                & precision                         & bf16 (fp32 master) \\
\bottomrule
\end{tabular}%
}
\end{table}

%% file: tex/appendix-proof.tex
\section{CAPR Search Efficiency}
\label{app:proof}

This appendix explains CAPR's compute advantage. The goal is not to prove a new convergence theorem, but to clarify why a block-level value head can recover much of the attribution signal that tree rollouts obtain by spending additional forward passes.

\subsection{Tree Search reduces Variance by spending Forwards}

Tree methods branch a partial denoising path, finish many continuations, and compare their final rewards. If a node has $B_{
m tree}^H$ leaves, the reward average at that node has variance roughly
\begin{equation}
\operatorname{Var}(\hat R_{\rm tree}\mid s_t) \approx \frac{\sigma^2_{R\mid s_t}}{B_{\rm tree}^H},
\end{equation}
where $\sigma^2_{R\mid s_t}$ is the outcome-reward variance after state $s_t$. This is useful, but the variance reduction is purchased by evaluating many continuations.

\subsection{CAPR learns the same Signal Once}

CAPR keeps only a small number of continuations, but it records the path state at every block boundary. For block boundary state $s_b$, the value head learns
\begin{equation}
V_\phi(s_b) \approx \mathbb{E}[R\mid s_b],
\end{equation}
so it can estimate which blocks made the final reward more likely before seeing many extra leaves. The TD residual
\begin{equation}
\delta_b = r_b + \gamma V_{\bar\phi}(s_{b+1}) - V_{\bar\phi}(s_b)
\end{equation}
turns one final reward into one block-level signal. This is the same role that a tree plays: it tells the optimiser which parts of a path were useful. The difference is that CAPR learns this mapping and reuses it, instead of rebuilding a fresh tree for every prompt.

\subsection{Why this is Stable}

Two safeguards keep the learned signal from dominating when it is unreliable. First, the target value head $V_{\bar\phi}$ is a slow Polyak average of the online head, which smooths noisy block targets. Second, CAPR gates the TD blend using the recent value loss: when the value head is poorly calibrated, $\beta_{\rm TD}^{\rm eff}$ becomes small and the update falls back toward the flat GRPO advantage.

The self-distillation loss and chi-square regulariser are local to the policy update and are defined in Appendix~\ref{app:stability} and Appendix~\ref{app:trust}. Self-distillation copies solved rollouts back into the LoRA policy, which is why the value head and search machinery can be removed at inference.

\subsection{Implication of the Bound}

CAPR trades tree depth for a learned block predictor. When the value head explains a nonzero part of the final-reward variance, the resulting block advantage has lower noise than a flat rollout advantage at nearly the same generation cost. When the value head is weak, the calibration gate reduces its influence, so CAPR reverts toward the flat baseline rather than relying on an inaccurate local signal.

This yields the following practical comparison:
\begin{equation}
\begin{aligned}
\text{tree search} &: \text{ lower variance by more leaves},\\
\text{CAPR} &: \text{ lower variance by learned block values}.
\end{aligned}
\end{equation}
The empirical diagnostics in \S\ref{sec:exp:dynamics} check this prediction. The value loss decreases, the $\chi^2$ statistic stays bounded, and the full CAPR run has the smoothest gradient-norm trajectory.

%% file: tex/appendix-baselines.tex
\section{Detailed Descriptions of dLLM RL Baselines}
\label{app:baselines}

This appendix summarises the baselines in Table~\ref{tab:main}. For each method, we describe its training signal, the advantage assigned to token updates, and the main approximation used for masked diffusion likelihoods. The goal is to make the comparison with CAPR interpretable rather than to reproduce the original derivations.

\paragraph{Diffu-GRPO~\citep{zhao2025d1}.}
A direct port of GRPO to masked dLLMs. The reward is the verifier outcome on the final completion $o_i$, and the advantage $A_{i,k} = r_i - \mathrm{mean}\{r_j\}_{j=1}^G$ is broadcast to every token of $o_i$. The only dLLM-specific piece is the log-probability estimator: rather than marginalising over denoising orders, Diffu-GRPO evaluates $\log\pi_\theta$ in a single forward pass on a randomly masked prompt and a fully masked completion, and re-samples the prompt mask each inner update for regularisation. This makes training cheap but gives every token the same rollout-level signal. CAPR instead uses the per-block advantages of \S\ref{sec:blockcritic}.

\paragraph{VRPO~\citep{zhu2025llada1_5}.}
A DPO-style preference optimiser using offline pairs $(q, o^w, o^l)$ and no explicit reward model. Since $\log\pi_\theta(o\mid q)$ is intractable, VRPO substitutes an ELBO bound $B_\pi(o\mid q) \le \log\pi(o\mid q)$ and forms the DPO score from differences of $B_{\pi_\theta} - B_{\pi_{\rm ref}}$ on $o^w$ and $o^l$. Two ideas reduce the variance of this estimator: an optimal allocation of the sampling budget across diffusion timesteps (more $n_t$, fewer $n_{z_t}$) and antithetic sampling that shares noise between the live and reference models. The remaining gap is structural: the score is built on a lower bound, not the true log-likelihood. VRPO is the recipe behind the public LLaDA-1.5 release on LLaDA-8B-Instruct.

\paragraph{wd1~\citep{tang2025wd1}.}
A ratio-free alternative to GRPO. After scoring a group with the verifier, wd1 maps the group-relative advantages $A_i = R_i - \mathrm{mean}(R_{1:G})$ to softmax weights $w_i^{\pm} = \mathrm{softmax}(\pm\psi A_i)$ and optimises $\sum_i(-w_i^{+} + w_i^{-})\log\pi_\theta(o_i\mid q)$, sidestepping the dLLM ratio entirely. The token-level log-probabilities are still approximated by the mean-field, single-pass estimator of Diffu-GRPO, with the same rollout-level signal for every token.

\paragraph{SAPO~\citep{xie2025step}.}
An augmented Diffu-GRPO with a step-aware reward. SAPO estimates whether an intermediate diffusion state makes progress by comparing the expected outcome of rollouts continued from two intermediate states. In practice, the cheaper variant continues from one intermediate state and the fully masked state. The combined signal $A_i^{\rm total} = A_i + \mathbf{1}[A_i\!>\!0]\,R_{\rm process}$ is again broadcast to all tokens, so the per-step granularity ends at the rollout level. Like Diffu-GRPO, SAPO inherits the mean-field likelihood approximation under randomly masked conditioning.

\paragraph{GDPO~\citep{rojas2025improving}.}
A GRPO variant that keeps the broadcast advantage but lifts the importance ratio from token-level (mean-field) to sequence-level via an ELBO surrogate, $r_i = L_{\rm ELBO}(o_i\mid q,\theta) / L_{\rm ELBO}(o_i\mid q,\theta_{\rm old})$. Variance is controlled by replacing the outer Monte Carlo over diffusion time with a small set of deterministic quadrature points (typically $N{=}2$ or $3$) and a lightweight inner Monte Carlo over masked states. As with VRPO, the optimisation runs on a bound rather than on the true log-likelihood. We include GDPO only in the LLaDA-MoE-7B-A1B-Instruct section of Table~\ref{tab:main} because~\citet{pan2025dtreerpo} report it only on that backbone under their shared evaluation setting.

\paragraph{TraceRL~\citep{wang2025revolutionizing}.}
A trajectory-aware PPO that optimises along the model's actual decoding path rather than under random masking. For each prompt, a rollout exposes a chain of intermediate states $(z_1,\ldots,z_0)$. For a token $o_j$ revealed at the transition $z_t \to z_{t-1}$, the importance ratio is computed on $\pi(\cdot\mid z_t)$. A diffusion-based value model $V_{\theta_v}$ then produces per-token GAE-style advantages, giving a fine-grained learning signal. The trade-off is the standard one: a learned value model is sensitive to value mis-specification and reward hacking, and the trajectory-conditioned likelihoods do not correct for the marginalisation over alternative decoding orders that lead to the same $o$.

\paragraph{d-TreeRPO~\citep{pan2025dtreerpo}.}
A tree-rollout baseline. Each prompt expands into a balanced tree of branching factor $B$ and depth $H$. Every node represents an $N/H$-step diffusion segment, and the per-tree-node advantage is averaged bottom-up from leaf rewards. This gives the finest non-CAPR advantage signal in our comparison, but the cost grows as $\tfrac{B(B^H-1)}{B-1}\!\cdot\!\tfrac{N}{H}$ forwards per prompt. CAPR targets similar attribution granularity with a single flat rollout structure, with cost linear in $G$ and independent of tree depth.

%% file: tex/appendix-reward-capr.tex
\section{Reward-Curve Comparison}
\label{app:reward-comparison}
\begin{figure*}[!t]
    \centering
    \includegraphics[width=0.6\linewidth]{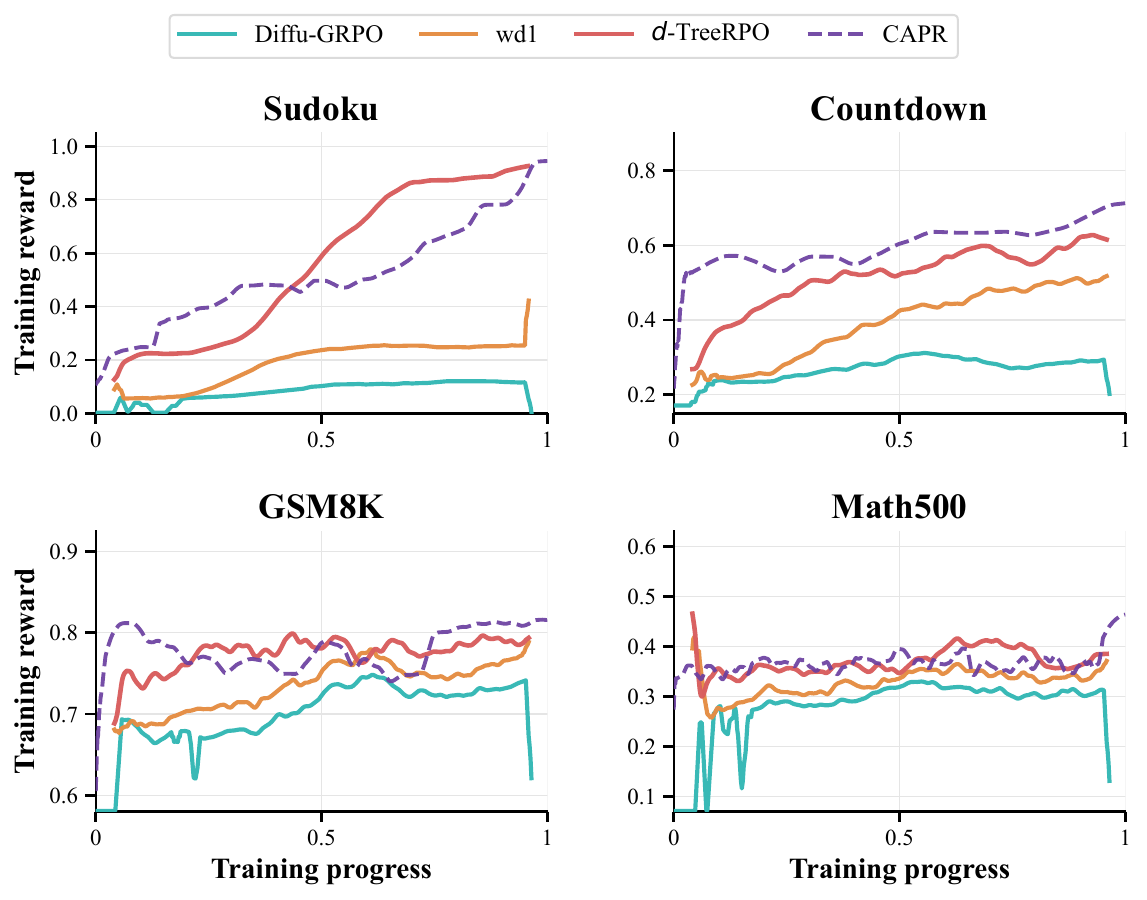}
    \caption{Training reward curves for CAPR compared with \emph{d}-TreeRPO and open-source baselines Diffu-GRPO and wd1 across the four reasoning tasks.}
    \label{fig:rewards-capr-dense}
\end{figure*}
Figure~\ref{fig:rewards-capr-dense} compares CAPR training reward curves with \emph{d}-TreeRPO and the open-source Diffu-GRPO~\citep{zhao2025d1} and wd1~\citep{tang2025wd1} baselines. CAPR moves from the low-reward region into the same high-reward band as the tree method while using the cheaper cached/branched rollout structure. On GSM8K and Math500, the base model already has stronger task competence and the reward axis is much narrower, so CAPR stays close to the \emph{d}-TreeRPO curve and may move slightly above or below it at different parts of training. This is the expected behaviour: on easier dense-backbone tasks, the main benefit is stable without paying for full tree expansion, rather than a large absolute change in training reward.

%% file: tex/appendix-prompt.tex
\section{Training Details}
\label{app:training-details}
\subsection{Prompts}
Prompts used in Sudoku, Countdown, GSM8k and Math500 are as follows:
\begin{MyBox}
### Sudoku

Please solve the following 4x4 Sudoku puzzle. The puzzle is provided as a 16-character string reading left-to-right, top-to-bottom, where '0' represents empty cells.

Rules:
- Fill empty cells with digits 1-4
- Each row must contain digits 1-4 exactly once
- Each column must contain digits 1-4 exactly once
- Each 2x2 box must contain digits 1-4 exactly once

Important: Your solution must be a COMPLETE 16-character string with only the digits 1-4, representing your final solved grid.

Respond in this exact format:
<reasoning>
Your step-by-step solving process
</reasoning>
<answer>
[16-character solution string with no spaces or separators]
</answer>

Solve the following Sudoku puzzle: {PUZZLE}
\end{MyBox}
\begin{MyBox}
### Countdown

Respond in the following format:
<reasoning>
...
</reasoning>
<answer>
...
</answer>

Using only the numbers {NUMBERS}, create an arithmetic expression that evaluates to exactly {TARGET}. You must use all numbers from the list, and each number must be used exactly once. You may use the operations +, -, *, and / as needed. After reasoning, provide only your final expression inside <answer></answer> tags without including an equals sign or the target number. For example, if the numbers are [2, 3, 4] and the target is 5, a valid answer is: <answer>2*4-3</answer>.
\end{MyBox}
\begin{MyBox}
### GSM8k

Respond in the following format:
<reasoning>
...
</reasoning>
<answer>
...
</answer>

You are a math expert. You will be given a question to solve. Solve it step by step. Put only the final answer (the number) in the <answer> </answer> tag without any other text.

{QUESTION}
\end{MyBox}
\begin{MyBox}
### Math500

Respond in the following format:
<reasoning>
...
</reasoning>
<answer>
...
</answer>

You are a math expert. You will be given a question to solve. Solve it step by step. Wrap the final answer in a \\boxed{{}}.

{QUESTION}
\end{MyBox}

\subsection{Reward Functions}
\noindent\textbf{Sudoku.} The reward is defined as the percentage of correctly filled previously empty cells, with a maximum value of 1 and a minimum value of 0.

\noindent\textbf{Countdown.} The reward is 1.0 if the provided expression achieves the target using the given numbers. If the numbers used are correct but the target is missed, the reward is 0.1. In all other cases, the reward is 0.

\noindent\textbf{GSM8k.} During training, we utilize a correctness-based reward, assigning a value of 1.0 if the provided answer matches the ground truth.

\noindent\textbf{Math500.} Similarly, during training, we employ a correctness-based reward system, assigning a reward of 1.0 if the given answer matches the ground truth.

\subsection{Dataset Links}
Following previous works~\citep{zhao2025d1,tang2025wd1}, we adopt the 4×4 Sudoku training and test sets provided by Diffu-GRPO~\citep{zhao2025d1} for Sudoku. For Countdown, models are trained on the 3to4 Countdown task training set\footnote{\url{https://huggingface.co/datasets/Jiayi-Pan/Countdown-Tasks-3to4}} and evaluated using the test set provided by Diffu-GRPO. For GSM8K\footnote{\url{https://huggingface.co/datasets/openai/gsm8k}} and Math500\footnote{train: \url{https://huggingface.co/datasets/ankner/math-500}, test: \url{https://huggingface.co/datasets/HuggingFaceH4/MATH-500}}, we strictly follow their official training-test splits for both training and evaluation. All datasets were used as per their intended use.

%% file: tex/appendix-case.tex
\section{Case Study}
\label{app:case}
This section presents a case study comparing CAPR against its strongest baseline, d-TreeRPO, on the same GSM8K question, as shown in Figure~\ref{fig:capr-vs-dtreerpo-case}. CAPR produces both a correct reasoning process and the correct final answer.

\begin{figure*}[t]
\begin{case}[title=LLaDA-8B-Instruct + CAPR, breakable=false]
\textbf{Question:} Digimon had its 20th anniversary. When it came out John was twice as old as Jim. If John is 28 now how old is Jim?

\vspace{3pt}

\textbf{Model Output:} 

\texttt{<reasoning>} 

To determine Jim's current age, we need to follow these steps:

1. Identify John's age when Digimon came out.

2. Determine Jim's age when Digimon came out.

3. Calculate Jim's current age.

First, we know that Digimon had its 20th anniversary, so it came out 20 years ago. If John is currently 28 years old, then his age when Digimon came out was $28 - 20 = 8$ years old.

Next, we know that when Digimon came out, John was twice as old as Jim. Therefore, let's set up the equation: $8 = 2 \cdot \text{Jim's age}$. To find Jim's age, we divide both sides by 2: $\text{Jim's age} = 8 / 2 = 4$.

Now, we know Jim was 4 years old when Digimon came out. Since Digimon came out 20 years ago, \textcolor{green!60!black}{Jim's current age is $4 + 20 = 24$ years old. \textcolor{green!60!black}{\ding{51}}}

\texttt{</reasoning>}

\texttt{<answer>}

$\boxed{24}$ \textcolor{green!60!black}{\ding{51}}

\texttt{</answer>}

\end{case}

\vspace{6pt}

\begin{case}[title=LLaDA-8B-Instruct + \emph{d}-TreeRPO, breakable=false]
\textbf{Question:} Digimon had its 20th anniversary. When it came out John was twice as old as Jim. If John is 28 now how old is Jim?

\vspace{3pt}

\textbf{Model Output:} 

\texttt{<reasoning>} 

To determine Jim's current age, we need to follow these steps:

1. Understand that Digimon had its 20th anniversary, 20 years ago.

2. At that time, John was twice as old as Jim.

3. We are currently given that John is 28 years old.

Let's denote Jim's current age as $J$.

Since Digimon had its 20th anniversary 20 years ago, John was $28 - 20 = 8$ years old at that time.

At that time, John was twice as old as Jim. Therefore, we can set up the equation: $8 = 2J$. To find Jim's current age, we solve for $J$: $J = \frac{8}{2} = 4$.

\textcolor{red}{Thus, Jim is currently 4 years old. \textcolor{deepred}{\ding{55}}}

\texttt{</reasoning>}

\texttt{<answer>}

$\boxed{4}$ \textcolor{deepred}{\ding{55}} \quad (\textit{ground truth:} $24$)

\texttt{</answer>}
\end{case}
\caption{A case study comparing LLaDA-8B-Instruct trained with \emph{CAPR} (top) and \emph{d}-TreeRPO (bottom) on the same GSM8K question. Both models correctly derive that Jim was 4 years old 20 years ago. \emph{CAPR} carries the temporal offset through the final step ($4 + 20 = 24$), whereas \emph{d}-TreeRPO conflates ``Jim's age then'' with ``Jim's current age $J$'' and stops at 4.}
\label{fig:capr-vs-dtreerpo-case}
\end{figure*}

%% file: custom.bib
@misc{pan2025dtreerpo,
      title={d-TreeRPO: Towards More Reliable Policy Optimization for Diffusion Language Models}, 
      author={Leyi Pan and Shuchang Tao and Yunpeng Zhai and Zheyu Fu and Liancheng Fang and Minghua He and Lingzhe Zhang and Zhaoyang Liu and Bolin Ding and Aiwei Liu and Lijie Wen},
      year={2026},
      eprint={2512.09675},
      archivePrefix={arXiv},
      primaryClass={cs.CL},
      url={https://arxiv.org/abs/2512.09675}, 
}

@misc{huang2025chipomythos,
      title={Correcting the Mythos of KL-Regularization: Direct Alignment without Overoptimization via Chi-Squared Preference Optimization}, 
      author={Audrey Huang and Wenhao Zhan and Tengyang Xie and Jason D. Lee and Wen Sun and Akshay Krishnamurthy and Dylan J. Foster},
      year={2025},
      eprint={2407.13399},
      archivePrefix={arXiv},
      primaryClass={cs.AI},
      url={https://arxiv.org/abs/2407.13399}, 
}

@inproceedings{cho2014learning,
    title = "Learning Phrase Representations using {RNN} Encoder{--}Decoder for Statistical Machine Translation",
    author = {Cho, Kyunghyun  and
      van Merri{\"e}nboer, Bart  and
      Gulcehre, Caglar  and
      Bahdanau, Dzmitry  and
      Bougares, Fethi  and
      Schwenk, Holger  and
      Bengio, Yoshua},
    editor = "Moschitti, Alessandro  and
      Pang, Bo  and
      Daelemans, Walter",
    booktitle = "Proceedings of the 2014 Conference on Empirical Methods in Natural Language Processing ({EMNLP})",
    month = oct,
    year = "2014",
    address = "Doha, Qatar",
    publisher = "Association for Computational Linguistics",
    url = "https://aclanthology.org/D14-1179/",
    doi = "10.3115/v1/D14-1179",
    pages = "1724--1734"
}

@misc{nie2025llada,
      title={Large Language Diffusion Models}, 
      author={Shen Nie and Fengqi Zhu and Zebin You and Xiaolu Zhang and Jingyang Ou and Jun Hu and Jun Zhou and Yankai Lin and Ji-Rong Wen and Chongxuan Li},
      year={2025},
      eprint={2502.09992},
      archivePrefix={arXiv},
      primaryClass={cs.CL},
      url={https://arxiv.org/abs/2502.09992}, 
}

@misc{zhu2025lladamoe,
      title={LLaDA-MoE: A Sparse MoE Diffusion Language Model}, 
      author={Fengqi Zhu and Zebin You and Yipeng Xing and Zenan Huang and Lin Liu and Yihong Zhuang and Guoshan Lu and Kangyu Wang and Xudong Wang and Lanning Wei and Hongrui Guo and Jiaqi Hu and Wentao Ye and Tieyuan Chen and Chenchen Li and Chengfu Tang and Haibo Feng and Jun Hu and Jun Zhou and Xiaolu Zhang and Zhenzhong Lan and Junbo Zhao and Da Zheng and Chongxuan Li and Jianguo Li and Ji-Rong Wen},
      year={2025},
      eprint={2509.24389},
      archivePrefix={arXiv},
      primaryClass={cs.CL},
      url={https://arxiv.org/abs/2509.24389}, 
}

@misc{zhu2025llada1_5,
      title={LLaDA 1.5: Variance-Reduced Preference Optimization for Large Language Diffusion Models}, 
      author={Fengqi Zhu and Rongzhen Wang and Shen Nie and Xiaolu Zhang and Chunwei Wu and Jun Hu and Jun Zhou and Jianfei Chen and Yankai Lin and Ji-Rong Wen and Chongxuan Li},
      year={2025},
      eprint={2505.19223},
      archivePrefix={arXiv},
      primaryClass={cs.LG},
      url={https://arxiv.org/abs/2505.19223}, 
}

@misc{cheng2025sdar,
      title={SDAR: A Synergistic Diffusion-AutoRegression Paradigm for Scalable Sequence Generation}, 
      author={Shuang Cheng and Yihan Bian and Dawei Liu and Linfeng Zhang and Qian Yao and Zhongbo Tian and Wenhai Wang and Qipeng Guo and Kai Chen and Biqing Qi and Bowen Zhou},
      year={2025},
      eprint={2510.06303},
      archivePrefix={arXiv},
      primaryClass={cs.LG},
      url={https://arxiv.org/abs/2510.06303}, 
}

@misc{wang2025revolutionizing,
      title={Revolutionizing Reinforcement Learning Framework for Diffusion Large Language Models}, 
      author={Yinjie Wang and Ling Yang and Bowen Li and Ye Tian and Ke Shen and Mengdi Wang},
      year={2025},
      eprint={2509.06949},
      archivePrefix={arXiv},
      primaryClass={cs.CL},
      url={https://arxiv.org/abs/2509.06949}, 
}

@misc{khanna2025mercury,
      title={Mercury: Ultra-Fast Language Models Based on Diffusion}, 
      author={Inception Labs and Samar Khanna and Siddhant Kharbanda and Shufan Li and Harshit Varma and Eric Wang and Sawyer Birnbaum and Ziyang Luo and Yanis Miraoui and Akash Palrecha and Stefano Ermon and Aditya Grover and Volodymyr Kuleshov},
      year={2025},
      eprint={2506.17298},
      archivePrefix={arXiv},
      primaryClass={cs.CL},
      url={https://arxiv.org/abs/2506.17298}, 
}

@misc{ye2025dream,
      title={Dream 7B: Diffusion Large Language Models}, 
      author={Jiacheng Ye and Zhihui Xie and Lin Zheng and Jiahui Gao and Zirui Wu and Xin Jiang and Zhenguo Li and Lingpeng Kong},
      year={2025},
      eprint={2508.15487},
      archivePrefix={arXiv},
      primaryClass={cs.CL},
      url={https://arxiv.org/abs/2508.15487}, 
}

@misc{song2025seed,
      title={Seed Diffusion: A Large-Scale Diffusion Language Model with High-Speed Inference}, 
      author={Yuxuan Song and Zheng Zhang and Cheng Luo and Pengyang Gao and Fan Xia and Hao Luo and Zheng Li and Yuehang Yang and Hongli Yu and Xingwei Qu and Yuwei Fu and Jing Su and Ge Zhang and Wenhao Huang and Mingxuan Wang and Lin Yan and Xiaoying Jia and Jingjing Liu and Wei-Ying Ma and Ya-Qin Zhang and Yonghui Wu and Hao Zhou},
      year={2025},
      eprint={2508.02193},
      archivePrefix={arXiv},
      primaryClass={cs.CL},
      url={https://arxiv.org/abs/2508.02193}, 
}

@inproceedings{sahoo2024simple,
  author       = {Subham S. Sahoo and
                  Marianne Arriola and
                  Yair Schiff and
                  Aaron Gokaslan and
                  Edgar Marroquin and
                  Justin T. Chiu and
                  Alexander Rush and
                  Volodymyr Kuleshov},
  editor       = {Amir Globersons and
                  Lester Mackey and
                  Danielle Belgrave and
                  Angela Fan and
                  Ulrich Paquet and
                  Jakub M. Tomczak and
                  Cheng Zhang},
  title        = {Simple and Effective Masked Diffusion Language Models},
  booktitle    = {Advances in Neural Information Processing Systems 38: Annual Conference
                  on Neural Information Processing Systems 2024, NeurIPS 2024, Vancouver,
                  BC, Canada, December 10 - 15, 2024},
  year         = {2024},
  url          = {http://papers.nips.cc/paper\_files/paper/2024/hash/eb0b13cc515724ab8015bc978fdde0ad-Abstract-Conference.html},
  bibsource    = {dblp computer science bibliography, https://dblp.org}
}

@inproceedings{shi2024simplified,
  author       = {Jiaxin Shi and
                  Kehang Han and
                  Zhe Wang and
                  Arnaud Doucet and
                  Michalis K. Titsias},
  editor       = {Amir Globersons and
                  Lester Mackey and
                  Danielle Belgrave and
                  Angela Fan and
                  Ulrich Paquet and
                  Jakub M. Tomczak and
                  Cheng Zhang},
  title        = {Simplified and Generalized Masked Diffusion for Discrete Data},
  booktitle    = {Advances in Neural Information Processing Systems 38: Annual Conference
                  on Neural Information Processing Systems 2024, NeurIPS 2024, Vancouver,
                  BC, Canada, December 10 - 15, 2024},
  year         = {2024},
  url          = {http://papers.nips.cc/paper\_files/paper/2024/hash/bad233b9849f019aead5e5cc60cef70f-Abstract-Conference.html},
  bibsource    = {dblp computer science bibliography, https://dblp.org}
}

@misc{nie2024scaling,
      title={Scaling up Masked Diffusion Models on Text}, 
      author={Shen Nie and Fengqi Zhu and Chao Du and Tianyu Pang and Qian Liu and Guangtao Zeng and Min Lin and Chongxuan Li},
      year={2025},
      eprint={2410.18514},
      archivePrefix={arXiv},
      primaryClass={cs.AI},
      url={https://arxiv.org/abs/2410.18514}, 
}

@misc{ye2023diffusion,
      title={Diffusion Language Models Can Perform Many Tasks with Scaling and Instruction-Finetuning}, 
      author={Jiasheng Ye and Zaixiang Zheng and Yu Bao and Lihua Qian and Quanquan Gu},
      year={2025},
      eprint={2308.12219},
      archivePrefix={arXiv},
      primaryClass={cs.CL},
      url={https://arxiv.org/abs/2308.12219}, 
}

@misc{zhao2025d1,
      title={d1: Scaling Reasoning in Diffusion Large Language Models via Reinforcement Learning}, 
      author={Siyan Zhao and Devaansh Gupta and Qinqing Zheng and Aditya Grover},
      year={2025},
      eprint={2504.12216},
      archivePrefix={arXiv},
      primaryClass={cs.CL},
      url={https://arxiv.org/abs/2504.12216}, 
}

@misc{tang2025wd1,
      title={wd1: Weighted Policy Optimization for Reasoning in Diffusion Language Models}, 
      author={Xiaohang Tang and Rares Dolga and Sangwoong Yoon and Ilija Bogunovic},
      year={2026},
      eprint={2507.08838},
      archivePrefix={arXiv},
      primaryClass={cs.LG},
      url={https://arxiv.org/abs/2507.08838}, 
}

@misc{wang2025d2,
      title={d2: Improved Techniques for Training Reasoning Diffusion Language Models}, 
      author={Guanghan Wang and Gilad Turok and Yair Schiff and Marianne Arriola and Volodymyr Kuleshov},
      year={2026},
      eprint={2509.21474},
      archivePrefix={arXiv},
      primaryClass={cs.LG},
      url={https://arxiv.org/abs/2509.21474}, 
}

@misc{xie2025step,
      title={Advancing Reasoning in Diffusion Language Models with Denoising Process Rewards}, 
      author={Shaoan Xie and Lingjing Kong and Xiangchen Song and Xinshuai Dong and Guangyi Chen and Eric P. Xing and Kun Zhang},
      year={2026},
      eprint={2510.01544},
      archivePrefix={arXiv},
      primaryClass={cs.AI},
      url={https://arxiv.org/abs/2510.01544}, 
}

@misc{yang2025taming,
      title={Taming Masked Diffusion Language Models via Consistency Trajectory Reinforcement Learning with Fewer Decoding Step}, 
      author={Jingyi Yang and Guanxu Chen and Xuhao Hu and Jing Shao},
      year={2025},
      eprint={2509.23924},
      archivePrefix={arXiv},
      primaryClass={cs.CL},
      url={https://arxiv.org/abs/2509.23924}, 
}

@misc{cobbe2021gsm8k,
      title={Training Verifiers to Solve Math Word Problems}, 
      author={Karl Cobbe and Vineet Kosaraju and Mohammad Bavarian and Mark Chen and Heewoo Jun and Lukasz Kaiser and Matthias Plappert and Jerry Tworek and Jacob Hilton and Reiichiro Nakano and Christopher Hesse and John Schulman},
      year={2021},
      eprint={2110.14168},
      archivePrefix={arXiv},
      primaryClass={cs.LG},
      url={https://arxiv.org/abs/2110.14168}, 
}

@inproceedings{lightman2023lets,
  author       = {Hunter Lightman and
                  Vineet Kosaraju and
                  Yuri Burda and
                  Harrison Edwards and
                  Bowen Baker and
                  Teddy Lee and
                  Jan Leike and
                  John Schulman and
                  Ilya Sutskever and
                  Karl Cobbe},
  title        = {Let's Verify Step by Step},
  booktitle    = {The Twelfth International Conference on Learning Representations,
                  {ICLR} 2024, Vienna, Austria, May 7-11, 2024},
  publisher    = {OpenReview.net},
  year         = {2024},
  url          = {https://openreview.net/forum?id=v8L0pN6EOi},
  bibsource    = {dblp computer science bibliography, https://dblp.org}
}

@misc{huang2025reinforcing,
      title={Reinforcing the Diffusion Chain of Lateral Thought with Diffusion Language Models}, 
      author={Zemin Huang and Zhiyang Chen and Zijun Wang and Tiancheng Li and Guo-Jun Qi},
      year={2025},
      eprint={2505.10446},
      archivePrefix={arXiv},
      primaryClass={cs.CL},
      url={https://arxiv.org/abs/2505.10446}, 
}

@misc{schulman2017proximal,
      title={Proximal Policy Optimization Algorithms}, 
      author={John Schulman and Filip Wolski and Prafulla Dhariwal and Alec Radford and Oleg Klimov},
      year={2017},
      eprint={1707.06347},
      archivePrefix={arXiv},
      primaryClass={cs.LG},
      url={https://arxiv.org/abs/1707.06347}, 
}

@misc{shao2024deepseekmath,
      title={DeepSeekMath: Pushing the Limits of Mathematical Reasoning in Open Language Models}, 
      author={Zhihong Shao and Peiyi Wang and Qihao Zhu and Runxin Xu and Junxiao Song and Xiao Bi and Haowei Zhang and Mingchuan Zhang and Y. K. Li and Y. Wu and Daya Guo},
      year={2024},
      eprint={2402.03300},
      archivePrefix={arXiv},
      primaryClass={cs.CL},
      url={https://arxiv.org/abs/2402.03300}, 
}

@misc{yu2025dapo,
      title={DAPO: An Open-Source LLM Reinforcement Learning System at Scale}, 
      author={Qiying Yu and Zheng Zhang and Ruofei Zhu and Yufeng Yuan and Xiaochen Zuo and Yu Yue and Weinan Dai and Tiantian Fan and Gaohong Liu and Lingjun Liu and Xin Liu and Haibin Lin and Zhiqi Lin and Bole Ma and Guangming Sheng and Yuxuan Tong and Chi Zhang and Mofan Zhang and Wang Zhang and Hang Zhu and Jinhua Zhu and Jiaze Chen and Jiangjie Chen and Chengyi Wang and Hongli Yu and Yuxuan Song and Xiangpeng Wei and Hao Zhou and Jingjing Liu and Wei-Ying Ma and Ya-Qin Zhang and Lin Yan and Mu Qiao and Yonghui Wu and Mingxuan Wang},
      year={2025},
      eprint={2503.14476},
      archivePrefix={arXiv},
      primaryClass={cs.LG},
      url={https://arxiv.org/abs/2503.14476}, 
}

@misc{zheng2025group,
      title={Group Sequence Policy Optimization}, 
      author={Chujie Zheng and Shixuan Liu and Mingze Li and Xiong-Hui Chen and Bowen Yu and Chang Gao and Kai Dang and Yuqiong Liu and Rui Men and An Yang and Jingren Zhou and Junyang Lin},
      year={2025},
      eprint={2507.18071},
      archivePrefix={arXiv},
      primaryClass={cs.LG},
      url={https://arxiv.org/abs/2507.18071}, 
}

@misc{cui2025entropy,
      title={The Entropy Mechanism of Reinforcement Learning for Reasoning Language Models}, 
      author={Ganqu Cui and Yuchen Zhang and Jiacheng Chen and Lifan Yuan and Zhi Wang and Yuxin Zuo and Haozhan Li and Yuchen Fan and Huayu Chen and Weize Chen and Zhiyuan Liu and Hao Peng and Lei Bai and Wanli Ouyang and Yu Cheng and Bowen Zhou and Ning Ding},
      year={2025},
      eprint={2505.22617},
      archivePrefix={arXiv},
      primaryClass={cs.LG},
      url={https://arxiv.org/abs/2505.22617}, 
}

@misc{li2025entropy,
      title={Entropy-Gated Branching for Efficient Test-Time Reasoning}, 
      author={Xianzhi Li and Ethan Callanan and Abdellah Ghassel and Xiaodan Zhu},
      year={2026},
      eprint={2503.21961},
      archivePrefix={arXiv},
      primaryClass={cs.CL},
      url={https://arxiv.org/abs/2503.21961}, 
}

@misc{zhang2024entropy,
      title={Entropy-Regularized Process Reward Model}, 
      author={Hanning Zhang and Pengcheng Wang and Shizhe Diao and Yong Lin and Rui Pan and Hanze Dong and Dylan Zhang and Pavlo Molchanov and Tong Zhang},
      year={2025},
      eprint={2412.11006},
      archivePrefix={arXiv},
      primaryClass={cs.LG},
      url={https://arxiv.org/abs/2412.11006}, 
}

@misc{tan2025gtpo,
      title={GTPO and GRPO-S: Token and Sequence-Level Reward Shaping with Policy Entropy}, 
      author={Hongze Tan and Zihan Wang and Jianfei Pan and Jinghao Lin and Hao Wang and Yifan Wu and Tao Chen and Zhihang Zheng and Zhihao Tang and Haihua Yang},
      year={2026},
      eprint={2508.04349},
      archivePrefix={arXiv},
      primaryClass={cs.CL},
      url={https://arxiv.org/abs/2508.04349}, 
}

@misc{yang2025mmada,
      title={MMaDA: Multimodal Large Diffusion Language Models}, 
      author={Ling Yang and Ye Tian and Bowen Li and Xinchen Zhang and Ke Shen and Yunhai Tong and Mengdi Wang},
      year={2025},
      eprint={2505.15809},
      archivePrefix={arXiv},
      primaryClass={cs.CV},
      url={https://arxiv.org/abs/2505.15809}, 
}

@misc{yang2025entropic,
      title={EntroPIC: Towards Stable Long-Term Training of LLMs via Entropy Stabilization with Proportional-Integral Control}, 
      author={Kai Yang and Xin Xu and Yangkun Chen and Weijie Liu and Jiafei Lyu and Zichuan Lin and Deheng Ye and Saiyong Yang},
      year={2026},
      eprint={2511.15248},
      archivePrefix={arXiv},
      primaryClass={cs.LG},
      url={https://arxiv.org/abs/2511.15248}, 
}

@misc{rojas2025improving,
      title={Improving Reasoning for Diffusion Language Models via Group Diffusion Policy Optimization}, 
      author={Kevin Rojas and Jiahe Lin and Kashif Rasul and Anderson Schneider and Yuriy Nevmyvaka and Molei Tao and Wei Deng},
      year={2026},
      eprint={2510.08554},
      archivePrefix={arXiv},
      primaryClass={cs.LG},
      url={https://arxiv.org/abs/2510.08554}, 
}

@misc{ou2025principled,
      title={Principled RL for Diffusion LLMs Emerges from a Sequence-Level Perspective}, 
      author={Jingyang Ou and Jiaqi Han and Minkai Xu and Shaoxuan Xu and Jianwen Xie and Stefano Ermon and Yi Wu and Chongxuan Li},
      year={2025},
      eprint={2512.03759},
      archivePrefix={arXiv},
      primaryClass={cs.CL},
      url={https://arxiv.org/abs/2512.03759}, 
}

@misc{lin2025boundary,
      title={Boundary-Guided Policy Optimization for Memory-efficient RL of Diffusion Large Language Models}, 
      author={Nianyi Lin and Jiajie Zhang and Lei Hou and Juanzi Li},
      year={2025},
      eprint={2510.11683},
      archivePrefix={arXiv},
      primaryClass={cs.LG},
      url={https://arxiv.org/abs/2510.11683}, 
}

@misc{he2025mdpo,
      title={MDPO: Overcoming the Training-Inference Divide of Masked Diffusion Language Models}, 
      author={Haoyu He and Katrin Renz and Yong Cao and Andreas Geiger},
      year={2025},
      eprint={2508.13148},
      archivePrefix={arXiv},
      primaryClass={cs.LG},
      url={https://arxiv.org/abs/2508.13148}, 
}

@misc{zhao2025inpainting,
      title={Inpainting-Guided Policy Optimization for Diffusion Large Language Models}, 
      author={Siyan Zhao and Mengchen Liu and Jing Huang and Miao Liu and Chenyu Wang and Bo Liu and Yuandong Tian and Guan Pang and Sean Bell and Aditya Grover and Feiyu Chen},
      year={2025},
      eprint={2509.10396},
      archivePrefix={arXiv},
      primaryClass={cs.LG},
      url={https://arxiv.org/abs/2509.10396}, 
}

@inproceedings{todorov2006linearly,
  author       = {Emanuel Todorov},
  editor       = {Bernhard Sch{\"{o}}lkopf and
                  John C. Platt and
                  Thomas Hofmann},
  title        = {Linearly-solvable Markov decision problems},
  booktitle    = {Advances in Neural Information Processing Systems 19, Proceedings
                  of the Twentieth Annual Conference on Neural Information Processing
                  Systems, Vancouver, British Columbia, Canada, December 4-7, 2006},
  pages        = {1369--1376},
  publisher    = {{MIT} Press},
  year         = {2006},
  url          = {https://proceedings.neurips.cc/paper/2006/hash/d806ca13ca3449af72a1ea5aedbed26a-Abstract.html},
  bibsource    = {dblp computer science bibliography, https://dblp.org}
}

@article{sutton1988learning,
  author       = {Richard S. Sutton},
  title        = {Learning to Predict by the Methods of Temporal Differences},
  journal      = {Mach. Learn.},
  volume       = {3},
  pages        = {9--44},
  year         = {1988},
  url          = {https://doi.org/10.1007/BF00115009},
  doi          = {10.1007/BF00115009},
  bibsource    = {dblp computer science bibliography, https://dblp.org}
}

@inproceedings{schulman2015high,
  author       = {John Schulman and
                  Philipp Moritz and
                  Sergey Levine and
                  Michael I. Jordan and
                  Pieter Abbeel},
  editor       = {Yoshua Bengio and
                  Yann LeCun},
  title        = {High-Dimensional Continuous Control Using Generalized Advantage Estimation},
  booktitle    = {4th International Conference on Learning Representations, {ICLR} 2016,
                  San Juan, Puerto Rico, May 2-4, 2016, Conference Track Proceedings},
  year         = {2016},
  url          = {http://arxiv.org/abs/1506.02438},
  bibsource    = {dblp computer science bibliography, https://dblp.org}
}

@inproceedings{kakade2002approximately,
  author       = {Sham M. Kakade and
                  John Langford},
  editor       = {Claude Sammut and
                  Achim G. Hoffmann},
  title        = {Approximately Optimal Approximate Reinforcement Learning},
  booktitle    = {Machine Learning, Proceedings of the Nineteenth International Conference
                  {(ICML} 2002), University of New South Wales, Sydney, Australia, July
                  8-12, 2002},
  pages        = {267--274},
  publisher    = {Morgan Kaufmann},
  year         = {2002},
  bibsource    = {dblp computer science bibliography, https://dblp.org}
}

@inproceedings{lillicrap2015continuous,
  author       = {Timothy P. Lillicrap and
                  Jonathan J. Hunt and
                  Alexander Pritzel and
                  Nicolas Heess and
                  Tom Erez and
                  Yuval Tassa and
                  David Silver and
                  Daan Wierstra},
  editor       = {Yoshua Bengio and
                  Yann LeCun},
  title        = {Continuous control with deep reinforcement learning},
  booktitle    = {4th International Conference on Learning Representations, {ICLR} 2016,
                  San Juan, Puerto Rico, May 2-4, 2016, Conference Track Proceedings},
  year         = {2016},
  url          = {http://arxiv.org/abs/1509.02971},
  bibsource    = {dblp computer science bibliography, https://dblp.org}
}
